\documentclass[fleqn,10pt]{wlscirep}
\pdfoutput=1
\usepackage[utf8]{inputenc}
\usepackage[T1]{fontenc}
\usepackage{subfig}
\usepackage{lineno}
\usepackage{float}
\usepackage{afterpage}
\usepackage{soul}
\usepackage{bbm}
\usepackage{tikz}
\usetikzlibrary{arrows.meta, positioning, fit, backgrounds}

\hypersetup{colorlinks=true, allcolors=black}

\usepackage{acronym}
\newacro{CLS}{causal local states}
\newacro{NGRC}{next-generation reservoir computing}
\newacro{RC}{reservoir computing}
\newacro{TE}{transfer entropy}
\newacro{CCM}{convergent cross mapping}
\newacro{MSE}{mean squared error}
\newacro{VPS}{valid prediction steps}
\newacro{VPT}{valid prediction time}
\newacro{TPR}{true positive rate}
\newacro{FPR}{false positive rate}
\newacro{UKPG}{UK power grid}
\newacro{L96}{Lorenz96}

\usepackage[resetlabels]{multibib}
\newcites{SI}{Supplementary References}

\newcounter{suppnote}
\newcommand{\suppnote}[1]{%
  \refstepcounter{suppnote}%
  \phantomsection
  \section*{Supplementary Note \thesuppnote: #1}%
}

\title{Causal Local States: Scalable Simultaneous Causal Network Inference and Forecasting for
Dynamical Systems}

\author[1,2]{Jonas Braun}
\author[3]{Fabian Fischbach}
\author[3,4]{Daniel K{\"o}glmayr}
\author[3,4]{Sebastian Baur}
\author[1,4,5*]{Christoph R{\"a}th}
\affil[1]{Ludwig-Maximilians-Universität München, Faculty of Physics, Geschwister-Scholl-Platz 1, 80539 Munich, Germany}
\affil[2]{Institut für Theoretische Physik, Universität zu Köln, Zülpicher Strasse 77, 50937 Cologne, Germany}

\affil[3]{Institut f{\"u}r KI-Sicherheit,
Deutsches Zentrum f{\"u}r Luft- und Raumfahrt (DLR), St. Augustin \& Ulm, Germany}
\affil[4]{Entrox Systems, Munich, Germany}
\affil[5]{Institut f{\"u}r Frontier Materials auf der Erde und im Weltraum, Deutsches Zentrum f{\"u}r Luft- und Raumfahrt (DLR), 51170 Cologne, Germany}

\affil[*]{christoph.raeth@dlr.de 
}

\begin{abstract}
Machine learning methods predict many real-world systems with remarkable accuracy, but they are typically treated as black boxes that offer no insight into which interactions drive the dynamics. Causal discovery methods reconstruct the interaction network from observational data, but without regard to whether the inferred structure supports prediction. Existing approaches combining both tasks rely on a single global hyperparameter, such as a causal threshold or a fixed neighborhood size, which cannot recover the structure of heterogeneous systems. Here we introduce causal local states (CLS), a framework that simultaneously infers an approximate Granger-causal interaction network and forecasts the system dynamics. For each node independently, we select the smallest set of neighbors that allows a predictive model to forecast the node near-optimally, and the resulting neighborhoods are then combined for a forecast of the full system. On three benchmarks of increasing difficulty, we achieve reconstruction of the underlying networks with high fidelity and forecasts on par with a model that is supplied with the true network, providing a step toward explainable and scalable forecasting of complex systems.
\end{abstract}

\begin{document}

\flushbottom
\maketitle
\thispagestyle{empty}

\section*{Introduction}

    Many of the systems that shape our lives, such as power grids~\cite{Witthaut2022}, the atmosphere~\cite{GhilLucarini2020} or ecosystems~\cite{Sugihara_CCM}, are high-dimensional, nonlinear dynamical systems. Anticipating how these systems evolve is essential for countless decisions, and machine learning methods can forecast complex dynamics with considerable success \cite{shahi2022prediction,kong2021machine}. However, these models are typically treated as black boxes: they turn raw, unstructured measurements into forecasts without revealing which of the many variables actually interact. Structuring the raw data into an interaction network during a prediction task helps in multiple ways. It allows treatment of high-dimensional systems where a lot of variables have little causal influence on the variable we want to forecast by selecting only a relevant subset of variables instead of considering the full system. Furthermore, an interaction network provides insights into the underlying system, supporting the decisions made from the data instead of blindly trusting a black-box approach. \Acf{CLS} achieves exactly that: recovering the interaction network from unstructured raw time series data and using that network to improve the system forecast.

In the following, we approach this problem by using ideas from two different fields. 
Causal discovery aims to reconstruct the interaction network of a nonlinear dynamical system from observational time series data, either by thresholding directed bivariate causal measures
such as Granger causality \cite{Granger1969}, \acf{TE} \cite{Schreiber00_TE}, and \acf{CCM} \cite{Sugihara_CCM},
or by multivariate algorithms that additionally address indirect links and confounding \cite{Causal_discovery_survey,Runge_causalRecon2018}.
By applying these methods, causal graphs can be reconstructed that are interpretable yet are inferred without taking into account their utility for forecasting the system's evolution. 
Data-driven forecasting methods, in particular \acf{RC} \cite{jaeger2004harnessing} and its computationally cheaper successor, \acf{NGRC} \cite{gauthier2021next,barbosa2022learning}, predict dynamical systems accurately at low training cost. Reservoir computers, however, process all variables of the time series simultaneously. While some structural insight may be recoverable in low-dimensional settings, for high-dimensional and unstructured data the trained model provides no reliable insight into the underlying interaction network or even becomes computationally infeasible to train and use.

\newpage
In this work, we develop a hybrid method that combines causal discovery and a prediction approach into one unified framework that (i) infers the interpretable interaction network of high-dimensional dynamical systems, in the form of one local neighborhood per variable, and (ii) uses these neighborhoods to produce accurate predictions of the whole system's temporal evolution. Because the interaction network is inferred explicitly rather than left implicit inside a monolithic model (a single model for the full system), it becomes a direct, inspectable output of the method alongside the forecast: for each variable, the framework reports the small set of other variables its prediction relies on. While we use \ac{NGRC} as the main predictive model, it is important to state that the causal measure and the model used are exchangeable components. This allows CLS to be adaptable to a wide range of systems.

The conceptual bridge between the two approaches is Granger's predictive definition of causality \cite{Granger1969}. A variable is said to Granger-cause another if its past improves the prediction of that other variable, provided it is conditioned on all other relevant data. Granger causality is thus predictive by construction. While this criterion is usually read as a causality test, it can equally be used to select between different variables. To facilitate prediction, a variable should only be passed to the model if and only if it improves the forecast. We achieve this by a wrapper-based approach. A black-box predictive model is trained and evaluated on candidate subsets of variables. The predictive performance across these subsets is then used to score the variables \cite{WrappersForFeatureSel,FeatureSelection}. Reconstructing an approximate Granger-causal interaction network therefore reduces to a variable-selection problem solved separately for each node, where variables are selected according to how much they improve the node's prediction accuracy. This identification of a wrapper-based  selection as an operational form of Granger causality is one of the central ideas behind this work. Furthermore, the selection of the right neighbors is not only relevant for the inferred structure but also for the forecast itself, because with finite data, processing irrelevant variables adds computational cost and degrades the model quality \cite{Trunk1979}.

For high-dimensional systems, knowing the interaction structure is not only desirable for interpretability but can also aid in making a prediction possible in the first place. To avoid the curse of dimensionality, scaling to higher-dimensional systems, Pathak et al. \cite{LocalStates} introduced a local-states approach, which decomposes a spatiotemporal system into fragments. Each fragment contains a core node that is predicted from a small local neighborhood by its own model. Baur and R\"ath\cite{Baur_2021} generalized this decomposition to systems with non-local interactions by constructing the neighborhoods from similarity measures such as cross-correlation and mutual information. Such decompositions make data-driven prediction of large systems feasible in the first place. However, a local state approach requires the data to be structured beforehand such that the interaction structure is known. Combining the previously described wrapper approach to structure the data and then subsequently using a local states approach for prediction allows us to predict a high-dimensional system while structuring the underlying data.

Only a small number of studies in the \ac{RC} literature connect network inference directly to predictive performance. Srinivasan et al. \cite{Parallel_ML_Srinivasan} \ combine \ac{TE} with a parallel \ac{RC} scheme, where links above a causal threshold are retained and the threshold is optimized as a global hyperparameter. Similarly, Chu et al.\ construct each neighborhood from the top-$q$ entries of a \ac{TE} matrix and again optimize the neighborhood size $q$ as a single global hyperparameter \cite{Global_q_inference_RCs}. These approaches assume that one global parameter, either a threshold or a neighborhood size, is appropriate for every node of the network. For heterogeneous systems this assumption fails, because scores are generally not comparable across subsystems with different dynamics and scales. Consequently, no global threshold or neighborhood size can recover all neighborhoods correctly, which we demonstrate explicitly in Supplementary Note~\ref{sec:supp_global}. A method in which every neighborhood is inferred independently, by a criterion local to its core node, is therefore still missing.  Further, Li et al.\cite{li2024higher} \ propose higher-order Granger reservoir computing, which evaluates candidate coupling configurations by their one-step prediction error and encodes the inferred structure directly into the reservoir architecture. While this approach seems very promising, it is unclear how the initial candidate set to start their algorithm with can be chosen efficiently for high-dimensional systems. With no prior system knowledge, the candidate set must include all candidate neighbors initially, which would explode the search space. \Ac{CLS} avoids this pitfall and allow a simultaneous network inference and prediction without any prior knowledge about the underlying network. The \ac{CLS} framework also does not require a global threshold, and since the per-node inference is fully parallel across nodes, it scales to high-dimensional systems naturally.

\section*{Results}
\begin{figure}
    \centering
    \includegraphics{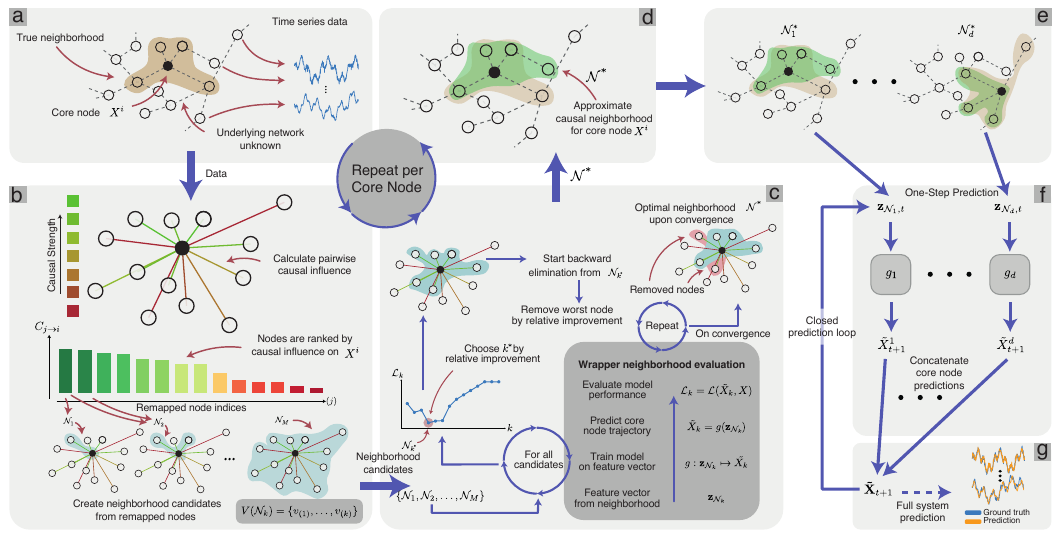}
    \caption{
Schematic illustration of the \acf{CLS} framework.
\textbf{a}, The input consists of high-dimensional time series data generated by a system with an unknown interaction network. 
\textbf{b}, A causal measure is used to rank candidate neighbors for a core node and generate neighborhood candidates. 
\textbf{c}, A wrapper-based optimization procedure selects the optimal neighborhood candidate and subsequently removes redundant features through backward elimination. 
\textbf{d}, The inference process yields an approximation of the causal neighborhood of the core node. 
\textbf{e}, After repeating \textbf{a-d} for all core nodes, the inferred local neighborhoods are combined into a global neighborhood matrix. 
\textbf{f}, The \ac{CLS} framework performs one-step prediction using the inferred neighborhoods by training a model $g_i$ on only the data present in the neighborhood $\mathcal{N}_i^*$.
\textbf{g}, One-step predictions are recursively fed back and concatenated to generate multi-step predictions of the full system dynamics. 
Due to the independent neighborhood inference and prediction procedure for each node, the \ac{CLS} framework is naturally parallelizable and scalable to high-dimensional systems. 
}
    \label{fig:Schematic}
\end{figure}
To establish our framework, we have to review some basic concepts central to \ac{CLS}. \Ac{CLS} aims to structure completely unstructured data prior to prediction. As the prediction then relies on a local-states approach \cite{LocalStates}, the structure \ac{CLS} aims to infer is the neighborhood matrix. The neighborhood matrix holds the neighborhood information for exactly one core node per row, whereas the core node just defines which specific node is in focus. Then the neighborhood is defined around a core node to contain all nodes that share a directed edge pointing towards this core node. Since each neighborhood is centered on a single core, $d$ neighborhoods cover the full system, and the collection can be stored compactly as a neighborhood matrix (a more strict definition can be found in Methods Eq. \ref{eq:neighborhood}). 

The first step of \ac{CLS} is to structure the unstructured high-dimensional input into neighborhoods for each core node. The algorithm infers the neighborhood for each core node separately, combines all the individual neighborhoods, and then uses the inferred network to forecast the whole system. The input is a collection of time series with unknown couplings (Fig.~\ref{fig:Schematic}a). For a given core node, a directed causal measure ranks all other variables as candidate neighbors and produces a small nested set of neighborhood candidates (Fig.~\ref{fig:Schematic}b). A wrapper model then selects the smallest candidate that predicts the core node near-optimally and prunes redundant members by backward elimination (Fig.~\ref{fig:Schematic}c), yielding an approximate Granger-causal neighborhood for that node (Fig.~\ref{fig:Schematic}d). Repeating this independently for every node assembles a global neighborhood matrix (Fig.~\ref{fig:Schematic}e). This matrix is then used to forecast the full system in parallel: each node is predicted from its own neighborhood alone (Fig.~\ref{fig:Schematic}f), and the one-step predictions are fed back recursively to generate the multi-step trajectory of the entire system (Fig.~\ref{fig:Schematic}g).

The justification for predicting a system through such local neighborhoods is the causal Markov condition. It guarantees that the parents of a variable are a sufficient input for predicting it: given its parents, a variable is conditionally independent of all its other non-descendants, so no further variable carries additional predictive information. This is what allows a single $d$-dimensional prediction problem to be replaced by $d$ small, local ones. Importantly, the condition is not violated by additional, spurious neighbors, because adding non-parents leaves the conditional independence intact. Spurious neighbors thus cost computation and obscure interpretation but do not invalidate the decomposition; \ac{CLS} nonetheless includes an elimination step to remove them, since smaller neighborhoods improve interpretability and simplify model fitting. One might assume that since a graph that fulfills the causal Markov condition is optimal for prediction, causal discovery algorithms would be preferable in network reconstruction over \ac{CLS}. We refer the reader to Supplementary Note~\ref{sec:supp_causal_vs_prediction} for a more thorough discussion on this matter.

We validate \ac{CLS} on three benchmarks of increasing difficulty.
On composite Lorenz63--R\"ossler systems of up to 90 dimensions, an exhaustive evaluation of all candidate neighborhoods confirms that the forecast error is a reliable selection criterion, and \ac{CLS} cleanly separates the constituent attractors in a regime where a single \ac{NGRC}, trained on the full system without variable selection, fails. On the 40-dimensional Lorenz96 system, \ac{CLS} recovers the banded coupling structure and produces forecasts comparable to those of a model supplied with the governing-equation adjacency matrix. Finally, on a higher-order Kuramoto model of the UK power grid, \ac{CLS} paired with a domain-informed wrapper reconstructs the real network topology almost perfectly and forecasts on par with a model given the true network. In all three cases, the inferred neighborhoods additionally provide insights into the system dynamics. \Ac{CLS} predicts the system and at the same time reports which interactions the prediction is based on.
\subsection*{Composite Lorenz63--R\"ossler attractor pairs}
\begin{figure}
    \centering
    \includegraphics{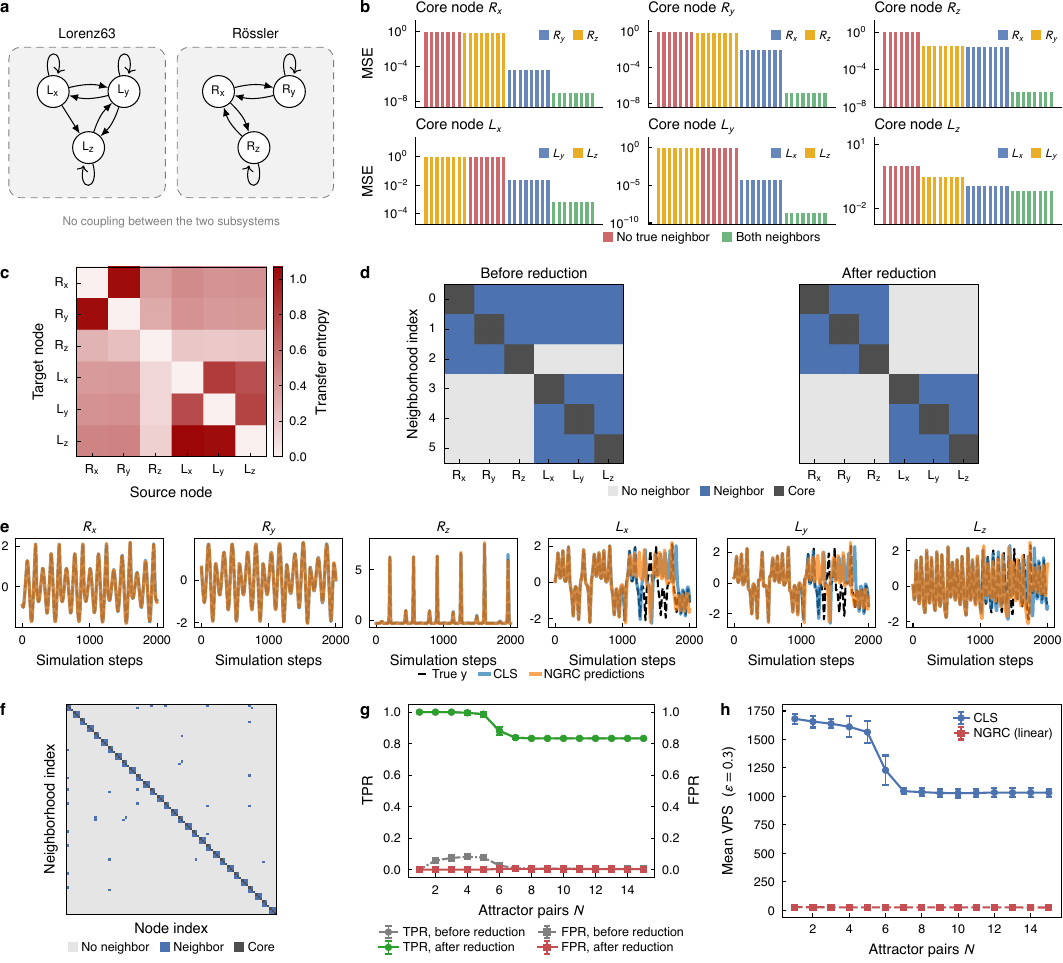}
    \caption{
    Analysis of composite (non-interacting) Lorenz63--Rössler systems using \acf{CLS}.
    \textbf{a}, Summary graph of a single Lorenz63--Rössler attractor pair. 
    \textbf{b}, Exhaustive wrapper validation for a single attractor pair. All non-empty neighborhood candidates of a core node are evaluated, and their wrapper \acf{MSE} is shown on a logarithmic axis, ordered by decreasing error. Bars are colored by how many of the core node's two same attractor neighbors the candidate contains: red, neither; blue and orange, exactly one (one color per neighbor, identified in the sub-panel legends); green, both.
    \textbf{c-e}, \Ac{CLS} analysis for a single attractor pair system.
    \textbf{c}, \Acf{TE} matrix computed during the filtering step.
    \textbf{d}, Inferred neighborhood matrices before and after feature elimination.
    \textbf{e}, Short-term prediction performance of \ac{CLS} compared with standard \acf{NGRC}.
    \textbf{f}, Inferred neighborhood matrix after feature elimination for a system of 15 coupled attractor pairs (90 variables), shown for seed 0.
    \textbf{g,h}, \Ac{CLS} performance across different system sizes ($N\in [1,15]$), using \ac{TE} as the filter and \ac{NGRC} as the wrapper model. Every system size was run for $n=10$ seeds, and every seed forecast from 10 disjoint starting points; markers give the mean over seeds and error bars $\pm 1$ standard deviation across them.
    \textbf{g}, Neighborhood inference accuracy: \acf{TPR} on the left axis (circles; grey dashed, before elimination; green solid, after) and \acf{FPR} on the right axis (squares; grey dash-dotted, before elimination; red solid, after).
    \textbf{h}, Prediction quality as the mean \acf{VPS} at $\epsilon = 0.3$, for \ac{CLS} (blue circles, solid line) and for a monolithic \ac{NGRC} trained on the full state vector with linear features only (red squares, dashed line).
    All experiment parameters can be extracted from Table \ref{tab:supp_l63r}.
    }
    \label{fig:lorenz_roessler}
\end{figure}

To test the \ac{CLS} framework on a well-tested system with a simple interaction structure,
we consider a dynamical system composed of $N$ non-interacting Lorenz63--R\"ossler attractor pairs (see Methods, \nameref{sec: systems}).
The two subsystems within each pair and distinct pairs share no coupling. Therefore, the interaction structure of a full $6N$-dimensional system decomposes cleanly into $2N$ independent three-dimensional blocks. A schematic of a single attractor pair is shown in Fig.~\ref{fig:lorenz_roessler}a. We emphasize that, despite this simple block structure, the system poses a non-trivial inference challenge, since a pairwise causal measure applied to time series data will in general return spurious causal relations between variables. In \ac{CLS} we therefore introduce an additional wrapper-elimination step designed to disentangle true from spurious neighbors (see Methods, \nameref{sec: CLS}).

\paragraph*{Validation of the wrapper model.} 
We first verify that a wrapper model can in principle identify the correct interaction structure when allowed to evaluate every possible interaction exhaustively. This is done for a system with one attractor pair $N=1$, as the space of possible interactions would be combinatorially huge otherwise.
For each core node $X_i$ we train a separate \ac{NGRC} model on each possible neighborhood candidate. Concretely, this means that an \ac{NGRC} is trained to forecast the core node trajectory exclusively from the data present in the neighborhood (Supplementary Note~\ref{sec:supp_wrapper}). Subsequently, we generate multiple one-step predictions and calculate the resulting \acf{MSE} from the true core node trajectory.

As shown in Fig.~\ref{fig:lorenz_roessler}b, the candidate neighborhoods that contain all true neighbors of the same attractor of the core node (green bars) yield one-step \ac{MSE}s several orders of magnitude lower than those that contain neither (red bars). Neighborhoods that contain one true neighbor but not the other tend to lie between these extremes (blue and orange bars, one color per neighbor). This monotone separation establishes the predictive loss as a reliable metric for neighborhood quality and motivates its use as the wrapper criterion in \ac{CLS}.
Exhaustive evaluation is, however, combinatorially infeasible beyond very small systems, as there are $2^{6N-1}-1$ non-empty neighborhood candidates per core node ($31$ for $N=1$). To restrict the search space before applying the wrapper evaluation, a filter step is introduced in the \ac{CLS} framework, generating a set of potential neighborhood candidates (described in more detail in Methods, \nameref{sec: CLS}).

\paragraph*{Inference and prediction for a single attractor pair.} 
We next apply the full \ac{CLS} framework to a single Lorenz63-R\"ossler pair ($N=1$) (see Methods, \nameref{sec: CLS}, for a thorough description of \ac{CLS}). Figure~\ref{fig:lorenz_roessler}c shows the \ac{TE} matrix computed during the filter step. Even though the block-diagonal structure is clearly visible, the cross entries are non-zero and, in places, comparable in magnitude to the weaker intra-attractor entries. For example, when inferring the neighborhood of $R_x$, the \ac{TE} assigned to its true neighbor $R_z$ is lower than the values assigned to the entirely independent Lorenz features $L_x, L_y, L_z$. A naive thresholding of the \ac{TE} matrix would, therefore, either miss true neighbors or admit false ones, depending on the threshold (Supplementary Note~\ref{sec:supp_global}). The filter step nevertheless performs its intended function of ranking the causal neighbors above the spurious ones for every core node, producing a small set of nested neighborhood candidates that the wrapper can evaluate efficiently. 
The feature elimination step then refines these candidates and removes the spurious and redundant features. Figure~\ref{fig:lorenz_roessler}d compares the neighborhood matrix obtained directly after the wrapper selects an initial candidate (left) with the matrix obtained after backward feature elimination (right). The initial candidate nearly captures the block-diagonal structure but retains several cross-attractor false positives. Backward elimination subsequently removes exactly these entries. The resulting reduced neighborhood matches the block-diagonal structure exactly. One needs to note that the block-diagonal structure just shows that \ac{CLS} successfully separates the attractors. According to Takens' embedding theorem~\cite{Takens_Theorem}, even dimensions that are not directly connected (e.g., $R_y$ and $R_z$) share information, so they can contribute to an increase in prediction performance. 

\paragraph*{Scaling to many attractor pairs.} 
To demonstrate that the per-node, parallel structure of \ac{CLS} scales to larger systems, we extend the experiment to $N=15$ attractor pairs, yielding a 90-dimensional system. Figure~\ref{fig:lorenz_roessler}f shows a sample inferred neighborhood matrix after wrapper elimination. The block-diagonal ground truth is recovered with high fidelity across all 15 blocks, with only a few false off-block entries.
Importantly, after computing the initial ranking using a bivariate causal measure, the inference for each core node is performed independently, so the computational complexity of \ac{CLS} scales linearly in the number of core nodes when parallelized. 
We systematically quantify inference quality and prediction performance as a function of system size by sweeping $N$ from 1 to 15. Figure~\ref{fig:lorenz_roessler}g reports the \acf{TPR} and \acf{FPR} of the inferred neighborhood matrices, both before and after the elimination step. After elimination, the \ac{TPR} remains above 0.8 across all system sizes tested, while the \ac{FPR} is suppressed to near zero.
Furthermore, Figure~\ref{fig:lorenz_roessler}h shows how these inference results translate into prediction performance, measured by the \acf{VPS} at error threshold $\epsilon = 0.3$ (Eq. \ref{eq:vps}). Standard \ac{NGRC} trained on the full state vector is not able to predict the system at all. Moreover, it was not possible to run \ac{NGRC} with higher-order nonlinearities (as was used to achieve the predictions in Figure ~\ref{fig:lorenz_roessler}e), as for a high-dimensional feature space, this requires too much RAM (see Methods, \nameref{sec: NGRC}). This limitation is lighter in \ac{CLS}, as \ac{CLS} uses a local-states approach following Pathak et al. \cite{LocalStates}, which keeps the feature space for each individual \ac{NGRC} instance small and manageable. The prediction performance of \ac{CLS} saturates at roughly 1000 \ac{VPS} for $N \gtrsim 7$.

These experiments establish two main points. First, the filter and wrapper steps of \ac{CLS} are complementary. The filter alone leaves cross-attractor false positives, while the wrapper elimination alone would be combinatorially infeasible and computationally expensive. Second, the per-node parallel architecture of \ac{CLS} allows for treatment of high-dimensional systems.

\subsection*{Lorenz96 system}
\begin{figure}
    \centering
    \includegraphics{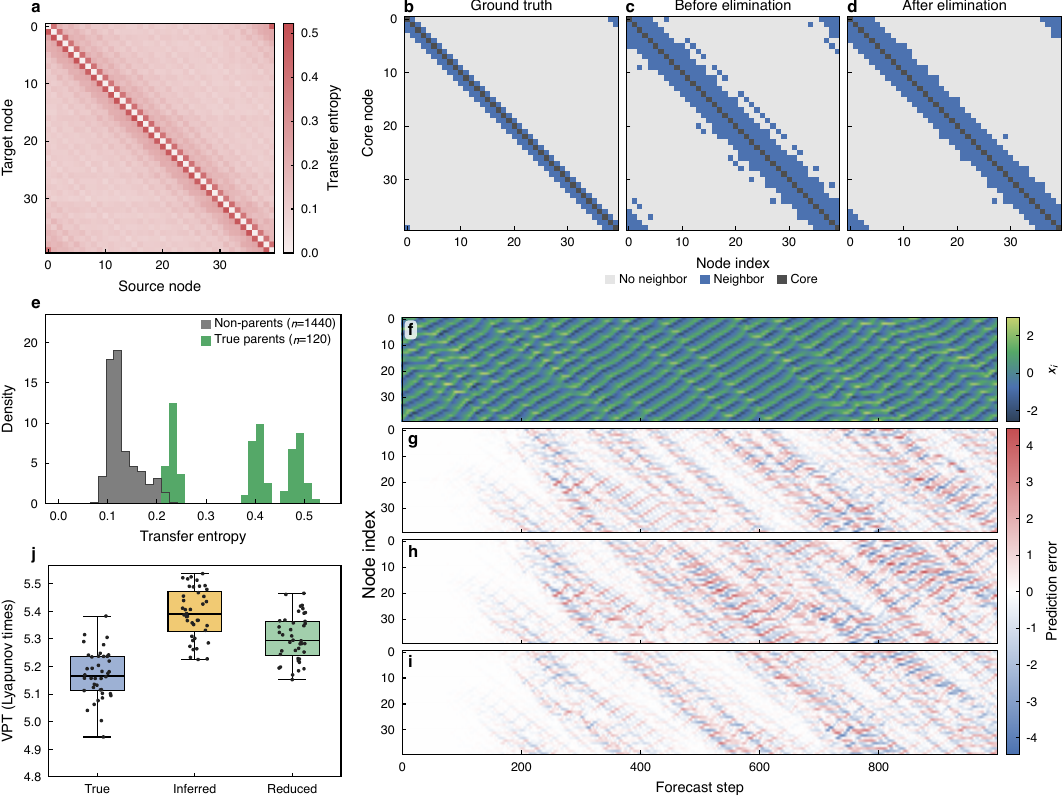}
    \caption{
        \Acf{CLS} analysis of a 40-dimensional Lorenz96 system.
        \textbf{a}, \Acf{TE} matrix computed during the filtering step.
        \textbf{b}, Ground-truth neighborhood matrix defined by the governing equations. 
        \textbf{c}, Inferred neighborhood matrix prior to feature elimination. 
        \textbf{d}, Final neighborhood matrix after feature elimination. 
        \textbf{e}, Histogram showing the distribution of calculated \ac{TE} values split between true parents of a node and non-parents.
        \textbf{f}, Ground-truth system trajectory. 
        \textbf{g--i}, Short-term predictions using the true adjacency matrix (\textbf{g}), the inferred matrix before feature elimination (\textbf{h}), and the inferred matrix after feature elimination (\textbf{i}). 
        \textbf{j},  \acf{VPS} at $\epsilon = 0.3$ expressed as \acf{VPT} in maximum Lyapunov times for \ac{CLS} using the three neighborhood matrices. Each of the 40 grid nodes contributes one point, its \ac{VPS} averaged over 100 forecasts (10 seeds $\times$ 10 disjoint launch points).
        All experiment parameters can be extracted from Table \ref{tab:supp_l96}.
        }
\label{fig:lorenz96}
\end{figure}

The composite attractor pairs in the previous section validate the wrapper criterion against an exhaustive ground truth. However, the subsystems in that benchmark are dynamically independent. We therefore consider the Lorenz96 system~\cite{Lorenz96}, a high-dimensional dynamic system that can exhibit chaotic behavior (see Methods, \nameref{sec: systems}).

\paragraph*{Structure inference.}
During the filter step, we compute the \ac{TE} matrix from the simulated trajectories (Fig.~\ref{fig:lorenz96}a). The resulting estimate reproduces the periodic coupling band of the ground-truth network, favoring the true parents of each core node (Fig.~\ref{fig:lorenz96}a, e). Applying wrapper selection and backward elimination with \ac{NGRC} as the wrapper model yields the neighborhood matrices shown in Fig.~\ref{fig:lorenz96}c,d. Both recover the underlying coupling pattern: across ten seeds every true parent is retained at both stages (\ac{TPR} $=1.000 \pm 0.000$), while backward elimination lowers the \ac{FPR} from $0.187 \pm 0.024$ to $0.146 \pm 0.007$. The inferred neighborhoods nevertheless contain more nodes than the three parents specified by the governing equations, averaging $9.75 \pm 0.87$ members before elimination and $8.25 \pm 0.23$ after. These additional nodes are not distributed randomly; they remain close to the core node on the lattice, with $93.2\%$ of the retained false positives lying within a lattice distance of four.

This outcome is consistent with a prediction-oriented selection criterion. Takens' embedding theorem~\cite{Takens_Theorem} implies that information relevant to a node's future state is not restricted to its direct dynamical parents. Nearby variables that are only indirectly connected can also encode predictive information. Including such nodes can therefore improve forecasts of the core variable, and a criterion based on predictive performance may retain them. Another possibility is that some of the extra connections reflect mild overfitting to the wrapper test set. As shown below, however, the retained nodes do not appear to degrade predictive performance.

\paragraph*{Prediction quality.}
To assess the impact of the inferred structure on forecasting performance, we train three otherwise identical \ac{NGRC} local-states models that differ only in their neighborhood matrix: the governing-equation adjacency, the inferred matrix before elimination, and the final matrix after elimination. Figure~\ref{fig:lorenz96}f--i compare the resulting short-term forecasts with the true trajectory, while Fig.~\ref{fig:lorenz96}j summarizes the \ac{VPT} across all nodes. All three models reproduce the short-term dynamics accurately. Averaged over ten seeds, the governing-equation adjacency reaches $5.18 \pm 0.67$ Lyapunov times, the inferred matrix before elimination $5.40 \pm 0.62$, and the reduced matrix after elimination $5.31 \pm 0.64$. \Ac{CLS} therefore matches the model supplied with the true adjacency matrix, despite working from a structure inferred entirely from data.

The interaction structure that yields the best predictions does not have to coincide with the governing-equation graph. The additional nearby connections retained by \ac{CLS} are absent from the nominal ground-truth network, yet they provide useful predictive information. Methods that treat the governing adjacency as the sole target of inference cannot capture this effect (Supplementary Note~\ref{sec:supp_causal_vs_prediction}). By selecting the neighborhoods according to forecast skill rather than causal graph recovery, \ac{CLS} directly optimizes the structure for predictions.


\subsection*{UK power grid system}

\begin{figure}
    \centering
    \includegraphics{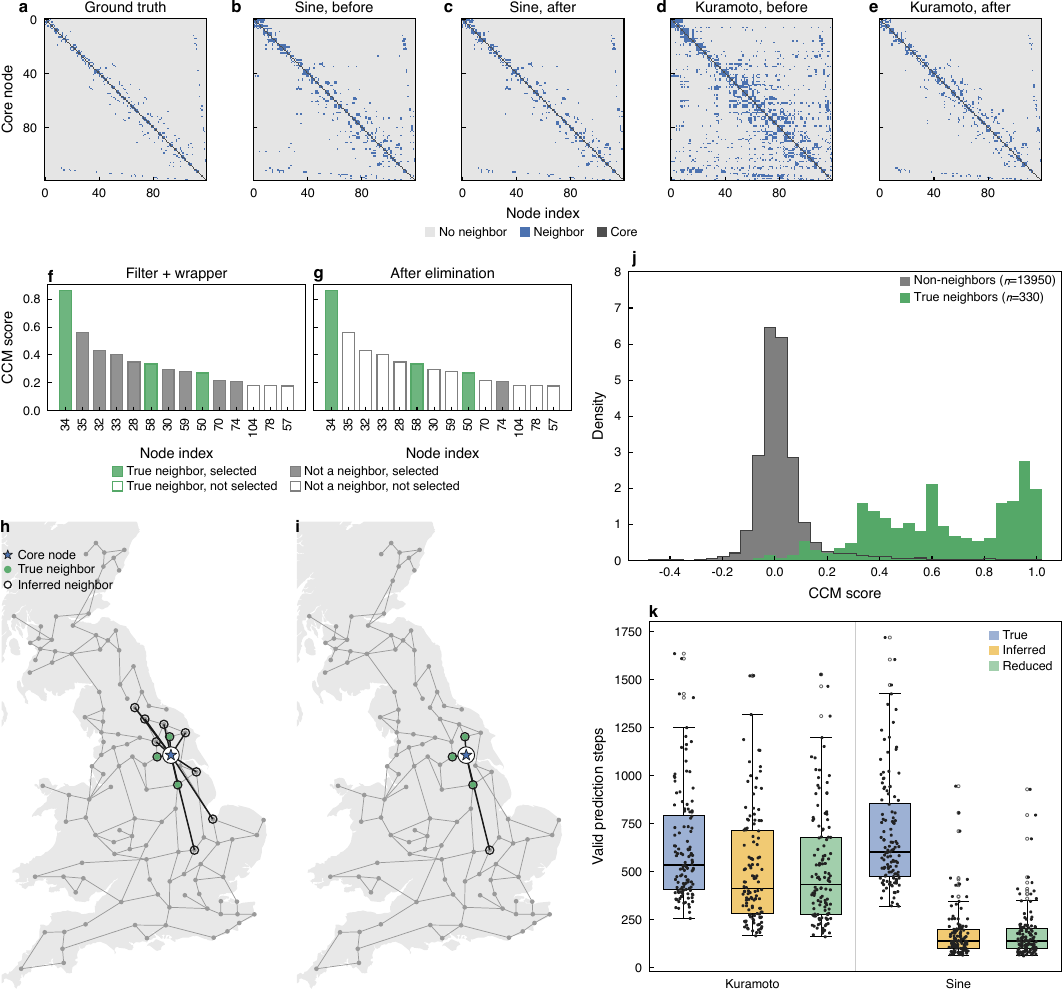}
    \caption{
        \Acf{CLS} analysis of the Kuramoto model of the UK power grid.
        \textbf{a}, Ground-truth adjacency matrix. 
        \textbf{b,c}, Neighborhood matrices inferred using Sine-NGRC before (\textbf{b}) and after (\textbf{c}) feature elimination.
        \textbf{d,e}, Neighborhood matrices inferred using Kuramoto-NGRC before (\textbf{d}) and after (\textbf{e}) feature elimination. 
        \textbf{f--i}, Example neighborhood inference process for a single core node using Kuramoto-NGRC as the wrapper model.
        \textbf{f, h}, Initial neighborhood identified after the filtering step. \textbf{f} shows that multiple spurious neighbors are selected in the first step because their \acf{CCM} score is higher than the true neighbors.
        \textbf{g, i}, Final neighborhood after backward feature elimination. The spurious neighbors are largely removed from the neighborhood.
        \textbf{j}, Histogram showing the distribution of calculated \ac{CCM} values split between true parents of a node and non-parents.
        \textbf{k}, \acf{VPS} for both wrapper models and all three neighborhood            matrices, on $\sin\theta$ at $\epsilon = 0.3$. Each of the 120 grid nodes            contributes one point, its \ac{VPS} averaged over 100 forecasts (10 seeds            $\times$ 10 disjoint launch points).
         All experiment parameters can be extracted from Table \ref{tab:supp_ukpg}.
                    }
    \label{fig:uk_grid}
\end{figure}

As a final and most challenging benchmark, we apply the \ac{CLS} framework to a real-world network, a dynamical model for the UK power grid in the form discussed by Li et al.\cite{li2024higher,li_2024_10685734} (see Methods, \nameref{sec: systems}).

This benchmark exposes a failure point of polynomial \ac{NGRC} on oscillator networks. As also reported in~\cite{Zhang2023} and consistent with our own experiments, the standard polynomial \ac{NGRC} formulation tends to produce divergent predictions on oscillator-type systems. As the \ac{CLS}
framework is model-agnostic, this allows us to sidestep this issue by selecting a wrapper whose feature dictionary is
better matched to the system at hand (Supplementary Note~\ref{sec:supp_wrapper}). We consider two such variants:

\begin{itemize}
    \item \textbf{Sine-NGRC.}  The nonlinear feature vector is built from polynomials of $\sin(\theta_i)$ and $\cos(\theta_i)$, rather than from the raw phases. This variant ensures that its features are bounded in $(-1,1)$, which naturally contains the divergence observed in polynomial \ac{NGRC}.
    \item \textbf{Kuramoto-NGRC.} The nonlinear expansion is built around the coupling nonlinearity of the Kuramoto model itself, using interaction terms of the form $\sin(\theta_j - \theta_i)$ for candidate node pairs $(i,j)$. This variant is more domain-informed and, like Sine-NGRC, has bounded features.
\end{itemize}
Both variants can serve as wrapper models within the \ac{CLS} framework. Before calculating \ac{CCM} \cite{Sugihara_CCM} to generate the variable ranking, we transform the data and apply \ac{CCM} on $\sin(\theta)$ instead of the raw $\theta$ values (this is necessary as \ac{CCM} relies on attractor reconstruction techniques that need a bounded trajectory). As the models transform the data with a sine-transformation during the nonlinear feature expansion, we pass the raw $\theta$ values. To improve stability, we do not choose $\theta_{t+1}$ as the training target for timestep $t$, but rather the difference from the current state $\Delta \theta_t=\theta_{t+1}-\theta_t$. The predicted $\theta_{t+1}$ value can then easily be calculated from $\Delta\theta_{t+1}$ 

We first illustrate the per-node inference procedure on a representative core node (node 51), whose true neighbors in the grid are nodes 34, 50 and 58 (Fig.~\ref{fig:uk_grid}f--i). The bar plot of pairwise causal influence calculated during the filter step (Fig.~\ref{fig:uk_grid}f) shows that several false positives have \ac{CCM} scores comparable to or even exceeding those of true neighbors, which makes them indistinguishable on the basis of the filter alone (Fig.~\ref{fig:uk_grid}j; Supplementary Note~\ref{sec:supp_global}). Thus, even though the candidate neighborhood returned by \ac{CLS} using Kuramoto-NGRC contains all true neighbors, it also includes a substantial number of spurious connections. The wrapper-based backward elimination step (Fig.~\ref{fig:uk_grid}g) then prunes these false positives: the remaining neighborhood almost perfectly matches the true local structure. 

We next apply the inference process to every node in the grid and assemble the inferred local neighborhoods into a global neighborhood matrix. 
Figures~\ref{fig:uk_grid}b--e show the neighborhood matrices inferred with both wrappers before and after backward elimination.
Two things stand out: First, for both wrapper variants, the elimination step substantially reduces the \ac{FPR} while largely preserving the \ac{TPR}. All values quoted here are means $\pm$ standard deviations over the ten seeds.
For Sine-NGRC the total \ac{FPR} drops from $2.57 \pm 0.25\%$ to $1.47 \pm 0.16\%$ with a \ac{TPR} falling from $86.4 \pm 2.4\%$ to $82.2 \pm 2.5\%$. Since the UK power grid is quite sparse, this is still favorable even though the percentage values might make it seem otherwise (Supplementary Note~\ref{sec:supp_causal_vs_prediction}).
For Kuramoto-NGRC, in comparison, the \ac{FPR} drops from $10.62 \pm 0.49\%$ to $1.87 \pm 0.15\%$ with the \ac{TPR} largely preserved ($98.1 \pm 0.5\%$ to $96.1 \pm 1.1\%$). With the neighborhood size bound at $d_{\max}=30$ for Kuramoto-NGRC, the \ac{CCM} filter places $98.8\%$ of all true neighbors among the $d_{\max}$ highest-ranked candidates. The \ac{TPR} of $98.1 \pm 0.5\%$ measured before elimination therefore sits essentially at this filter ceiling, which shows that the wrapper selection discards almost no true neighbors.

For each wrapper variant we run closed-loop forecasts using three different neighborhood matrices: the ground-truth adjacency, the initial (post-filter) inferred matrix, and the final (post-elimination) reduced matrix. 
The Kuramoto-NGRC predictions remain accurate over a long prediction horizon for all three neighborhood matrices, with both inferred ones yielding errors comparable to those obtained with the true graph.
The Sine-NGRC predictions reach a median $601$ \ac{VPS} against Kuramoto-NGRC's $535$. Sine only collapses on its own \emph{inferred} adjacencies ($601 \to 140$), which suggests that the failure is in the neighborhood selection.

Figure~\ref{fig:uk_grid}k summarizes the observations through the \ac{VPS} across all nodes computed on the $\sin(\theta)$ values. Most notably, however, the final prediction quality obtained by \ac{CLS} using Kuramoto-NGRC is comparable with a local-states approach that used the true underlying network.

\section*{Discussion}

We have introduced \acf{CLS}, a framework that simultaneously infers the interaction structure of a high-dimensional dynamical system and forecasts its evolution. \ac{CLS} extends the local-states~\cite{LocalStates} and generalized local-states~\cite{Baur_2021} approaches by removing their reliance on a known network structure or simple correlation-based measures. Where generalized local states use similarity measures to group variables, \ac{CLS} combines a filter- and wrapper-based approach to approximate each node's causal neighborhood directly from data. By bringing ideas from causal discovery into the prediction task, \ac{CLS} moves toward more explainable machine learning.

Across our benchmarks, \ac{CLS} was able to reconstruct the neighborhoods accurately and provide predictions on par with a local states approach that was provided the true network. For the coupled Lorenz63--R\"ossler task, \ac{CLS} was able to separate the different attractors and forecast all of them accurately. Applied to the Lorenz96 system, it recovered the banded interaction structure and matched the forecasting performance of a local-states model supplied with the true adjacency matrix. Paired with a domain-informed Kuramoto-NGRC model as a wrapper, \ac{CLS} reconstructed an almost exact adjacency matrix for the UK power grid and achieved short-term accuracy comparable to a model given the true network. Nonetheless, this experiment showed that model selection matters for performance, as \ac{CLS} with a Sine-NGRC wrapper did not generate adequate results on the UK power grid.
The independence of causal measure and wrapper model in the \ac{CLS} framework is also a lever here. By choosing a filter and wrapper that match the characteristics of the data, the performance of the algorithm can be greatly improved.

One further aspect of \ac{CLS} is worth emphasizing. Each neighborhood is inferred separately from the full system, so both inference and prediction parallelize across nodes and scale to high-dimensional systems.

Despite these promising results, several open challenges remain. Most notably, \ac{CLS} has not yet been applied to real-world datasets. Potential applications include financial time series, power grid dynamics, or weather forecasting, domains where both predictive performance and interpretability are of high importance.

In its current form, the inference procedure is restricted to identifying causal neighbors. Extending the framework to include the selection of relevant temporal features could provide deeper insights into system dynamics and further improve predictive performance. Incorporating tree-based \ac{NGRC} models with built-in feature selection capabilities could be a promising direction in this context \cite{ExtraTreesNGRC}. 

Another challenge concerns hyperparameter optimization. The \ac{CLS} framework involves multiple local models, each of which may require its own set of hyperparameters. The current approach is to use a shared hyperparameter configuration across all neighborhoods. However, this is unlikely to be optimal. Developing efficient strategies to optimize hyperparameters without exhaustive search remains an open challenge.

\Ac{CLS} predicts high-dimensional dynamical systems at scale while inferring an approximation of their underlying causal structure. The benchmark performance demonstrates the strong potential of \ac{CLS} as a tool for combining prediction and interpretability in complex systems.
\section*{Methods}

One of the core concepts used extensively in causal local states are neighborhoods.
Consider a multivariate time series $\{X_t\}_{t\in\mathbb{Z}}$ with $X_t = (X_t^1,\dots,X_t^d)$, generated by a dynamical system whose variables interact. Under the assumption of stationarity, the interaction network can be represented compactly by a summary graph, a directed graph whose nodes correspond to the system variables and which contains an edge $X^j \to X^i$ if and only if $X^j$ causally influences $X^i$ at some time lag~\cite{Summary_Graphs}. The summary graph abstracts away the lag structure of causal influence, which matches the scope of CLS: it performs spatial feature selection (which variables matter for a node) rather than temporal feature selection (which time lags matter).

For a core node $X^i$, we define its neighborhood as
\begin{equation}
\label{eq:neighborhood}
\mathcal{N}(X^i)
=
\bigl(
    \{X^i\} \cup \mathrm{Pa}(X^i),\;
    \{ (X^k, X^i) : X^k \in \mathrm{Pa}(X^i) \}
\bigr) \ ,
\end{equation}
where $\mathrm{Pa}(X^i)$ denotes the parent set of $X^i$ in the summary graph, i.e., all variables with a directed edge into $X^i$. The definition is close to Pearl's notion of a family~\cite{Pearl_2009} but distinguishes explicitly between the core node and its parents, and is adapted from the generalized local-states approach~\cite{Baur_2021}. 
\phantomsection
\subsection*{Causal Local States}
\makeatletter\def\@currentlabelname{Causal Local States}\makeatother
\label{sec: CLS}
The \ac{CLS} framework fragments a $d$-dimensional system into $d$ core nodes, each with a local neighborhood that is inferred directly from the data. Unlike the local-states approach~\cite{LocalStates}, \ac{CLS} treats the interaction structure as unknown and recovers it through a three-step procedure: causal filtering, wrapper selection, and backward elimination. The individually inferred neighborhoods are then combined into a global neighborhood matrix and used for prediction. Because the inference is run independently for each core node, the framework is trivially parallelizable and scales to high-dimensional systems. A full schematic is shown in Figure \ref{fig:Schematic} and the pseudocode is provided in Algorithm~\ref{alg:cls}.

Let $\{X_t\}_{t\in\mathbb{Z}}$ with $X_t = (X_t^1,\dots,X_t^d)$ be the observed time series, and let $X^j = \{X_t^j\}_{t\in\mathbb{Z}}$ denote the trajectory of node $j$. We describe the inference of the neighborhood $\mathcal{N}^*(X^i)$ for a single, fixed core node $X^i$, but the procedure is identical for all nodes. 

\paragraph{Filter step.}

An exhaustive search over the $2^{d-1}$ possible neighborhoods of a core node is infeasible in high dimensions, so we begin by ranking candidates using a bivariate causal measure. With this step, we restrict the search to a nested sequence of the highest-scoring candidates. For each $j\neq i$ we compute a causal score
\begin{equation}
\label{eq:cls_score}
C_{j\to i} := C(X^j \to X^i),
\end{equation}
where $C(\cdot\to\cdot)$ can be any directed causal measure (we use \ac{TE} and \ac{CCM} throughout, but the choice is interchangeable). Unlike symmetric statistical measures, a causal score captures directed influence. Reindexing the potential neighboring nodes such that $C_{(1)\to i}\geq C_{(2)\to i}\geq\dots$ yields a nested family of candidate neighborhoods of increasing size, defined as
\begin{equation}
\label{eq:cls_neighborhood_sequence}
\mathcal{N}_k (X^i) = \mathcal{N}_k=
\Bigl(
\{X^i\}\cup\{X^{(1)},\dots,X^{(k)}\},\;
\{(X^{(\ell)},X^i)\}_{\ell=1}^{k}
\Bigr),
\qquad 1 \le k \le d_{\max},
\end{equation}
where $X^{(k)}$ is the node with the $k$-th highest score and $d_{\max}$ is a hyperparameter bounding the neighborhood size. The filter step thus reduces the search space from $\mathcal{O}(2^d)$ to $d_{\max}+1$ candidates $\mathcal{C}_i = \{\mathcal{N}_k (X^i)\}_{k=1}^{d_{\max}}$ per core node. Calculating the causal matrix is in practice done prior to the separation between the core nodes and has itself a computational complexity of $\mathcal{O}(d^2)$. Depending on the wrapper used, $k=0$ is undefined or the neighborhood containing just the core node itself.  

\paragraph{Wrapper selection.}

Each neighborhood candidate is evaluated by a wrapper model~\cite{WrappersForFeatureSel}. A wrapper model is a concept from feature selection, where a black-box model is trained and evaluated on a given feature set to assess the feature quality. In \ac{CLS}, the wrapper is trained on the full data present in the neighborhood but scored only on the prediction of the core node. For the candidate, $\mathcal{N}_k$ we form a feature vector $\mathbf{z}_{k} = (X^j)_{X^j\in V(\mathcal{N}_k)}$, where the core node data $X^i$ may be excluded depending on the wrapper. Using a wrapper that supports an \ac{NGRC} inference-like mode is preferable. During inference mode the \ac{NGRC} model learns the mapping from the neighbors to the core node, which means that the neighbor contributions are not masked by the self-influence of the core node (see supplementary information~\ref{sec:supp_wrapper} for a closer look at the self-influence during neighborhood selection). Now we train a model $g_k:\mathbb{R}^{d_k}\to\mathbb{R}$ to predict the next core state, $\tilde{X}^i_{k,t+1} = g_k(\mathbf{z}_{k,t})$. The wrapper test loss $\mathcal{L}_k$ is evaluated on a wrapper test set $\mathcal{I}^{\mathrm{w\text{-}test}}$ disjoint from the wrapper training set. We control the size of the wrapper test set with the wrapper test fraction parameter, which just sets the fraction of the training data that is held out for wrapper evaluation. This is done so the final test set is completely unseen during the evaluation of the full system prediction. Finally, we select the smallest neighborhood that achieves near-optimal performance using a relative improvement criterion with threshold $\alpha_1\in(0,1)$. The selection scans candidates in order of increasing size and a candidate $\mathcal{N}_k$ becomes the new baseline $k^*$ whenever
\begin{equation}
\Delta(k,k^*) = \frac{\mathcal{L}_{k^*}-\mathcal{L}_{k}}{\mathcal{L}_{k^*}}\;\ge\; \alpha_1 .
\end{equation}

\paragraph{Backward elimination.}

The selected neighborhood $\mathcal{N}_{k^*}$ may still contain redundant variables. Starting from $\mathcal{N}=\mathcal{N}_{k^*}$ with a vertex set $V$ and edge set $E$ and a baseline loss $\mathcal{L}_0=\mathcal{L}_{k^*}$, we iteratively remove the least informative neighbor. For each neighbor node of $X^i$, $X^j\in\mathcal{N}$, $j\neq i$, the reduced neighborhood $\mathcal{N}^{(-j)} = (V\setminus\{X^j\},\, E\setminus\{(X^j,X^i)\})$ is re-evaluated to give a loss $\mathcal{L}_{-j}$. The candidate $X^m$ with $m=\arg\min_j\mathcal{L}_{-j}$, i.e.\ the node whose removal least degrades prediction, is eliminated only if the relative deterioration stays within a threshold $\alpha_2\in(0,1)$,
\begin{equation}
\frac{\mathcal{L}_{-m}-\mathcal{L}_0}{\mathcal{L}_0} \;\le\; \alpha_2 .
\end{equation}
The baseline is then updated and the step repeated until no further removal satisfies the criterion. The causal Markov condition motivates this procedure: non-descendants of $X^i$ are conditionally independent of $X^i$ given its parents, so removing a truly causal neighbor degrades prediction sharply, whereas removing a spurious one does not. The surviving variables define the approximate causal neighborhood $\mathcal{N}^*(X^i)$. The values of $d_{\max}$, $\alpha_1$ and $\alpha_2$ used for each benchmark are listed in Supplementary Tables~\ref{tab:supp_l63r}--\ref{tab:supp_ukpg}.

\paragraph{Full-system prediction.}

Repeating the inference for all core nodes yields neighborhoods $\{\mathcal{N}^*(X^i)\}_{i=1}^d$, which we store compactly as a
neighborhood matrix
\begin{equation}
A_{ij} =
\begin{cases}
2, & j = i,\\
1, & X^j \in V(\mathcal{N}^*(X^i)),\\
0, & \text{otherwise}.
\end{cases}
\end{equation}
Using this matrix, the full system is forecast with the local-states
approach: a model $g_i$ is trained for each node on its neighborhood
alone, one-step predictions are concatenated across nodes, and the
result is fed back recursively to generate multi-step forecasts of the
full system (Fig.~\ref{fig:Schematic}f,g).

\begin{algorithm}[t]
\caption{Causal local states: neighborhood inference for core node $X^i$}
\label{alg:cls}
\begin{algorithmic}[1]
\Require Time series $\mathbf{X}=(X^1,\dots,X^d)$, core index $i$,
causal measure $C$, wrapper loss $\mathcal{L}$, bounds $d_{\max}$,
thresholds $\alpha_1,\alpha_2$
\Ensure Approximate causal neighborhood $\mathcal{N}^*(X^i)$
\Statex \textit{// Filter step}
\For{$j=1$ to $d$, $j\neq i$}
    \State $C_{j\to i} \gets C(X^j\to X^i)$
\EndFor
\State Reindex so that $C_{(1)\to i}\ge\dots\ge C_{(d-1)\to i}$
\State $\mathcal{C}_i \gets \{\mathcal{N}_k\}_{k=0}^{d_{\max}}$ via
Eq.~\eqref{eq:cls_neighborhood_sequence}
\Statex \textit{// Wrapper selection}
\State $k^*\gets 0$ \Comment{or start at $1$ if wrapper requires a minimum size}
\For{$k=1$ to $d_{\max}$}
    \State train $g_k$ on $\mathcal{N}_k$; evaluate $\mathcal{L}_k$
    \If{$(\mathcal{L}_{k^*}-\mathcal{L}_k)/\mathcal{L}_{k^*}\ge\alpha_1$}
        \State $k^*\gets k$
    \EndIf
\EndFor
\Statex \textit{// Backward elimination}
\State $\mathcal{N}\gets\mathcal{N}_{k^*}$; \;$\mathcal{L}_0\gets\mathcal{L}(\mathcal{N})$
\Repeat
    \ForAll{$X^j\in V(\mathcal{N})$, $j\neq i$}
        \State $\mathcal{L}_{-j}\gets\mathcal{L}(\mathcal{N}^{(-j)})$
    \EndFor
    \State $m\gets\arg\min_j\mathcal{L}_{-j}$
    \If{$(\mathcal{L}_{-m}-\mathcal{L}_0)/\mathcal{L}_0\le\alpha_2$}
        \State $\mathcal{N}\gets\mathcal{N}^{(-m)}$; \;$\mathcal{L}_0\gets\mathcal{L}_{-m}$
    \Else
        \State \textbf{break}
    \EndIf
\Until{no candidate satisfies the removal criterion}
\State \Return $\mathcal{N}^*\gets\mathcal{N}$
\end{algorithmic}
\end{algorithm}
\phantomsection
\subsection*{Next-Generation Reservoir Computing}
\makeatletter\def\@currentlabelname{Next-Generation Reservoir Computing}\makeatother
\label{sec: NGRC}
\Ac{NGRC}, introduced by Gauthier et al.~\cite{gauthier2021next}, is the predictive model we use throughout this work, both as the wrapper in the inference steps and as the per-node model in the full-system forecast. Unlike classical \ac{RC}, \ac{NGRC} has no random recurrent reservoir. Its feature vector is instead built explicitly from the input data through time-delay coordinates and polynomial nonlinearities. This design is motivated by the equivalence between \ac{RC} and nonlinear vector autoregression~\cite{EquivalenceRC_NVAR}, and it reaches comparable forecast skill with less training data and fewer hyperparameters than a classical reservoir.

Let $\mathbf{X}_t \in \mathbb{R}^d$ be the input state at discrete time $t$. \Ac{NGRC} forms a feature vector $\mathbb{O}_t$ by concatenating a constant bias $c$, a linear part, and a nonlinear part,
\begin{equation}
\mathbb{O}_t = c \oplus \mathbb{O}_{\mathrm{lin},t} \oplus \mathbb{O}_{\mathrm{nl},t},
\end{equation}
where $\oplus$ denotes concatenation. The linear part is a time-delay embedding of the input,
\begin{equation}
\mathbb{O}_{\mathrm{lin},t} = \bigoplus_{i=1}^{k} \mathbf{X}_{t-(i-1)s} \;\in\; \mathbb{R}^{dk},
\end{equation}
with $k$ delays spaced $s$ steps apart. The nonlinear part collects the unique monomials of the linear features. For a polynomial order $p$ we write $\mathbb{O}_{\mathrm{nl},t}^{(p)} = \lceil \bigotimes_{i=1}^{p}\mathbb{O}_{\mathrm{lin},t}\rceil$, where $\lceil\cdot\rceil$ keeps one copy of each distinct monomial, and concatenate over a chosen set of orders $\mathcal{P}$,
\begin{equation}
\mathbb{O}_{\mathrm{nl},t} = \bigoplus_{p\in\mathcal{P}} \mathbb{O}_{\mathrm{nl},t}^{(p)}.
\end{equation}
The delays $k$, the spacing $s$, and the order set $\mathcal{P}$ are the hyperparameters of the feature map. As in classical \ac{RC}, only the linear readout is trained,
\begin{equation}
\tilde{\mathbf{Y}}_t = \mathbf{W}_{\mathrm{out}}\,\mathbb{O}_t,
\end{equation}
with $\mathbf{W}_{\mathrm{out}}$ obtained by ridge regression with regularization $\beta$. Multi-step forecasts are produced in closed loop by feeding the prediction back as the next input. The values of $k$, $s$, $\mathcal{P}$ and $\beta$ used for each benchmark, separately for the wrapper and for the local-states forecast, are listed in Supplementary Tables~\ref{tab:supp_l63r}--\ref{tab:supp_ukpg}.

\paragraph{Feature-space size.}
The nonlinear expansion gives \ac{NGRC} its accuracy, but it also limits the method on high-dimensional systems as the feature space grows in size combinatorially. A monomial of order $p$ corresponds to drawing $p$ factors from the $dk$ linear components with replacement and without regard to order, so the order-$p$ block contributes $\binom{p+dk-1}{p}$ features, and the full nonlinear feature vector has size
\begin{equation}
\label{eq:ngrc_feature_size}
\bigl|\mathbb{O}_{\mathrm{nl},t}\bigr| = \sum_{p\in\mathcal{P}} \binom{p+dk-1}{p}.
\end{equation}
A monolithic \ac{NGRC} model over the full state quickly produces feature vectors and design matrices too large to hold in memory. Restricting each model to a small neighborhood keeps $d$ small and the full-system forecast tractable.

\paragraph{Inference mode.}
The wrapper model during network inference ideally does not predict the core node from its own past but only from its candidate neighbors. \Ac{NGRC} supports this directly through an inference mode that learns a mapping between disjoint sets of variables, $\mathbf{X}_t \in \mathbb{R}^q \mapsto \mathbf{Y}_t \in \mathbb{R}^p$, rather than forecasting the input forward. Because each prediction depends only on the current input and its delays, the time steps are independent and can be evaluated in parallel, which is well suited to the many one-step wrapper evaluations of the inference stage. Using inference mode in the wrapper also ensures that a candidate neighbor's contribution is judged on its own predictive value and is not masked by the self-influence of the core node (Supplementary Note~\ref{sec:supp_wrapper}).

\subsection*{Causality measures}
The filter step of \ac{CLS} (Eq.~\eqref{eq:cls_score}) admits any directed, bivariate causality measure as a scoring function. In principle, the filter step would also admit domain-specific scoring functions (e.g. physical distance), but this was not further researched. We briefly review two common bivariate causality measures that are used in this work.

\paragraph{Transfer entropy.}
Transfer entropy~\cite{Schreiber00_TE} is a model-free, information-theoretic generalization of Granger causality. It measures the reduction in uncertainty about the next state of $X$ gained from the past of $Y$, beyond the past of $X$,
\begin{equation}
\label{eq:transfer_entropy}
T_{Y\to X} = H\!\left(X_{n+1}\mid X_n^{(k)}\right)
- H\!\left(X_{n+1}\mid X_n^{(k)}, Y_n^{(l)}\right),
\end{equation}
where $X_n^{(k)}$ and $Y_n^{(l)}$ are delay vectors of orders $k$ and $l$, and $H$ is the Shannon entropy. We estimate it with a histogram-based estimator. 

\paragraph{Convergent cross mapping.}
Convergent cross mapping (CCM)\cite{Sugihara_CCM} targets systems in which cause and effect are dynamically inseparable, where Granger-type measures break down. Using Takens' delay-embedding theorem~\cite{Takens_Theorem}, it reconstructs the shadow manifold of each variable and tests whether states of $X$ can be recovered from the nearest-neighbor structure of $Y$. If $X$ causally drives $Y$, this cross-map skill, calculated as the correlation $\rho_{X\to Y}$ between true and cross-mapped values, increases and converges as the library length grows. Because we only compare causal influences across candidate neighbors, we hold the embedding dimension, delay, and library length fixed and use the cross-map skill at fixed library size as the score. We acknowledge that this is an unusual usage of CCM, as one would usually increase the library length and investigate convergence. However, since the filter step is not interested in causal discovery but rather ranking the variables relative to each other, the usual workflow behind CCM had to be adapted.
\subsection*{Prediction quality metrics}

Two metrics are used to quantify short-term predictive performance throughout this work: the mean squared error (MSE) and the valid prediction steps (VPS) with the associated valid prediction time (VPT).

\paragraph{Mean squared error.}
Let $\mathbf{X}_t \in \mathbb{R}^{d}$ denote the true system state at time step $t$ and $\tilde{\mathbf{X}}_t \in \mathbb{R}^{d}$ the corresponding prediction. The MSE over a prediction horizon of $N_{\mathrm{pred}}$ steps starting at $t_0$ is
\begin{equation}
\label{eq:mse}
\mathrm{MSE} = \frac{1}{N_{\mathrm{pred}}} \sum_{k=1}^{N_{\mathrm{pred}}}
\left\| \tilde{\mathbf{X}}_{t_0+k} - \mathbf{X}_{t_0+k} \right\|_2^2.
\end{equation}
The MSE also serves as the wrapper loss $\mathcal{L}$ in the neighborhood inference steps of CLS: there, it is evaluated on the predicted trajectory of the core node alone ($d=1$), so that the wrapper score reflects only the quality of the core-node prediction and is unaffected by the other variables in the candidate neighborhood.

\paragraph{Valid prediction steps and valid prediction time.}
While the MSE averages over a fixed horizon, the VPS measures how long a closed-loop forecast remains close to the true trajectory. Given a threshold $\epsilon > 0$, the VPS is defined as the number of time steps until the prediction error exceeds the threshold,
\begin{equation}
\label{eq:vps}
k_v = \min \left\{ k \in \{1, \dots, N_{\mathrm{pred}}\} \;\middle|\;
\left\| \tilde{\mathbf{X}}_{t_0 + k} - \mathbf{X}_{t_0 + k} \right\|_2 > \epsilon \right\},
\end{equation}
and is set to $N_{\mathrm{pred}}$ if the error remains below $\epsilon$ for the entire horizon. The corresponding valid prediction time is $t_v = k_v\,\Delta t$, with $\Delta t$ the sampling interval. Where indicated, we express $t_v$ in units of the maximal Lyapunov time $1/\lambda_{\max}$ of the system. The threshold $\epsilon$ used for each experiment is reported together with the corresponding results.

\phantomsection
\subsection*{Benchmark systems}
\makeatletter\def\@currentlabelname{Benchmark systems}\makeatother
\label{sec: systems}
We validate \ac{CLS} on three systems of increasing difficulty. All parameter values are provided in tables \ref{tab:supp_l63r}, \ref{tab:supp_l96}, \ref{tab:supp_ukpg} in the supplementary information.

\paragraph{Composite Lorenz63--R\"ossler attractor pairs.}
The first benchmark is a system composed of $N$ non-interacting Lorenz63--R\"ossler attractor pairs. Each pair consists of a three-dimensional Lorenz63 subsystem \cite{lorenz1963deterministic},
\begin{equation}
\dot{L}_x = \sigma(L_y - L_x)\ , \qquad
\dot{L}_y = L_x(\rho - L_z) - L_y\ , \qquad
\dot{L}_z = L_x L_y - \beta L_z\ ,
\end{equation}
and a three-dimensional R\"ossler subsystem with state variables $(R_x, R_y, R_z)$ \cite{roessler1976},
\begin{equation}
\dot{R}_x = -R_y - R_z\ , \qquad \ \
\dot{R}_y = R_x + a R_y\ , \qquad \qquad \ \
\dot{R}_z = b + R_z(R_x - c)\ .
\end{equation}
The two subsystems within a pair share no dynamical coupling, and distinct pairs are likewise non-interacting, so the full $6N$-dimensional system decomposes exactly into $2N$ independent three-dimensional blocks with a block-diagonal ground-truth interaction network.

\paragraph{Lorenz96.}
The Lorenz96 (L96) system~\cite{Lorenz96} consists of $N$ variables arranged on a one-dimensional periodic lattice and evolving according to
\begin{equation}
\dot{x}_i(t) = \bigl(x_{i+1}(t) - x_{i-2}(t)\bigr)\,x_{i-1}(t) - x_i(t) + F,
\qquad i = 1,\dots,N,
\end{equation}
with cyclic indices $x_{i+N} = x_i$ and a constant forcing $F$ that controls the degree of chaos. This produces a banded, periodic ground-truth neighborhood matrix. We study a chaotic regime with $N=40$ and $F=5$, where we find a largest Lyapunov exponent of $\lambda_{max}\approx 0.42$~\cite{L96_LLE}.

\paragraph{Higher-order Kuramoto model of the UK power grid.}
The third benchmark is a dynamical model of the UK power grid in the form discussed by Li et al.~\cite{li2024higher,li_2024_10685734}, based on the topology of the UK high-voltage power transmission grid~\cite{Simonsen2008, Rohden2012}, comprising 120 nodes connected by 165 undirected edges. The grid dynamics generalize the Kuramoto model~\cite{kuramoto1975} to a higher-order network~\cite{skardal2020higher,Battiston.2026},
\begin{equation}
\label{eq:ukpg}
\dot{\theta}_i(t) = \omega_i + \gamma_1 \sum_{j} A_{ij}\left[\sin(\theta_j - \theta_i - \alpha) + \sin\alpha\right] + \gamma_2 \sum_{(j,k)} B_{ijk}\left[\sin(\theta_j + \theta_k - 2\theta_i - \beta) + \sin\beta\right]
\quad i = 1,\dots,N,
\end{equation}
where $\theta_i(t)$ and $\omega_i$ denote the phase and natural frequency of oscillator $i$. The coefficients $\gamma_1$ and $\gamma_2$ control pairwise and three-node interactions, respectively, while $A_{ij}$ is the adjacency matrix and $B_{ijk}$ the triple-coupling support tensor encoding the higher-order couplings. For some coupling regimes, such higher-order systems exhibit extremely complex dynamics \cite{skardal2020higher}. The natural frequencies are drawn once per run as $\omega_i = s_i a_i$, with absolute values $a_i \sim \mathcal{U}[\omega_{\mathrm{abs,low}},\, \omega_{\mathrm{abs,low}} + 1]$ and signs $s_i = \pm 1$ with equal probability, independently across oscillators. With $\omega_{\mathrm{abs,low}} = 0.3$ the resulting distribution is symmetric about zero but excludes a band around it, so no oscillator is close to stationary. The entries of $B_{ijk}$ are either $0$ or $1$, and there are six higher-order triangles in the grid: $B_{ijk} = 1$, for $\{14, 111, 112\}$, $\{33, 30, 32\}$, $\{49, 57, 114\}$, $\{81, 82, 117\}$, $\{84, 87, 88\}$ and $\{100, 95, 109\}$. For details of the parameters of the model, see Tab.~\ref{tab:supp_ukpg} and Li et al. ~\cite{li2024higher,li_2024_10685734}.

\section*{Author contributions statement}
J. B. developed the algorithm and performed the experimental studies. 
F. F. and S. B. helped with conceptualization of the initial algorithm and coding.  
C. R. initiated and supervised the work. All authors interpreted the findings and wrote the manuscript.

\section*{Data and Code Availability}
The data and code are available from the corresponding author upon reasonable request.

\section*{Acknowledgments}
We acknowledge helpful discussions with Max Mynter and Haochun Ma regarding causality measures.

\section*{Declaration of Interests}
The German Aerospace Center has filed a patent application related to this work.
C. R., S. B. and D. K. have financial interests as co-founders of Entrox Systems, which is commercializing RC. The remaining authors declare no competing interests.

\bibliography{sample}
\label{MainLastPage}

\appendix
\renewcommand{\thesection}{S\arabic{section}}

\clearpage
\rfoot{\small\sffamily\bfseries\thepage/\pageref{LastPage}}
\resetlinenumber
\section*{Supplementary Information}
\suppressfloats[t]
\setcounter{page}{1}
\setcounter{figure}{0}
\setcounter{table}{0}
\setcounter{equation}{0}

\captionsetup[figure]{name=Supplementary Figure}
\captionsetup[table]{name=Supplementary Table}
\acresetall

\makeatletter
\suppnote{A global causal threshold cannot reconstruct even a simple network}
\label{sec:supp_global}
\makeatother
\definecolor{corered}{HTML}{C44E52}
\definecolor{tehot}{HTML}{9C0606}
\newcommand{\drawsystem}[4]{%
\begin{tikzpicture}[
    every node/.style={circle, draw, minimum size=4.2mm, inner sep=0.5pt, font=\scriptsize, fill=white},
    title/.style={draw=none, fill=none, font=\scriptsize, inner sep=1pt},
    core/.style={circle, draw, thick, fill=corered!35},
    trueedge/.style={-{Latex}, semithick, shorten >=0.5pt, shorten <=0.5pt},
    spurious/.style={-{Latex}, dotted, semithick, shorten >=0.5pt, shorten <=0.5pt},
    systembox/.style={shape=rectangle, draw=gray!45, fill=gray!10, rounded corners=2pt,
                      dashed, line width=0.4pt, inner sep=2.5pt},
    baseline=(Lx.south)
]
    \ifthenelse{\equal{#2}{Lx}}
    {\node[core] (Lx) {$L_x$};}
    {\node (Lx) {$L_x$};}
    \ifthenelse{\equal{#2}{Ly}}
    {\node[core] (Ly) [below left=0.50cm and 0.02cm of Lx] {$L_y$};}
    {\node (Ly) [below left=0.50cm and 0.02cm of Lx] {$L_y$};}
    \ifthenelse{\equal{#2}{Lz}}
    {\node[core] (Lz) [below right=0.50cm and 0.02cm of Lx] {$L_z$};}
    {\node (Lz) [below right=0.50cm and 0.02cm of Lx] {$L_z$};}
    \node (Ltitle) [title, above=0.14cm of Lx] {Lorenz63};
    \ifthenelse{\equal{#2}{Rx}}
    {\node[core] (Rx) [right=1.55cm of Lx] {$R_x$};}
    {\node (Rx) [right=1.55cm of Lx] {$R_x$};}
    \ifthenelse{\equal{#2}{Ry}}
    {\node[core] (Ry) [below left=0.50cm and 0.02cm of Rx] {$R_y$};}
    {\node (Ry) [below left=0.50cm and 0.02cm of Rx] {$R_y$};}
    \ifthenelse{\equal{#2}{Rz}}
    {\node[core] (Rz) [below right=0.50cm and 0.02cm of Rx] {$R_z$};}
    {\node (Rz) [below right=0.50cm and 0.02cm of Rx] {$R_z$};}
    \node (Rtitle) [title] at (Rx |- Ltitle) {#1};
    \foreach \from/\to in {#3} { \draw[trueedge] (\from) -- (\to); }
    \foreach \from/\to in {#4} { \draw[spurious] (\from) -- (\to); }
    \begin{scope}[on background layer]
        \node[systembox, fit=(Lx) (Ly) (Lz)] {};
        \node[systembox, fit=(Rx) (Ry) (Rz)] {};
    \end{scope}
\end{tikzpicture}%
}

The approaches of Srinivasan et al.~\citeSI{SI_Parallel_ML_Srinivasan} and Chu et al.~\citeSI{SI_Global_q_inference_RCs} determine every neighborhood of the network with a single global hyperparameter. Srinivasan et al. propose a setup where they calculate a causal score for all system links. Then only those links that exceed a global threshold are retained, with the global threshold being a hyperparameter that can be optimized. Chu et al. follow a similar approach where instead of selecting by absolute causal value, each core node's neighborhood is built by the $q$ highest-scoring candidates. The neighborhood size can then also be optimized as a global hyperparameter. Here we demonstrate, on the composite Lorenz63--R\"ossler system, why neither strategy can recover the structure of a heterogeneous system and, thus more generally, why the neighborhood criterion must instead be local to each core node.

\begin{figure}[t]
\centering
\begin{minipage}[t]{0.46\linewidth}
    \centering
    {\bfseries a}\par\vspace{0.5em}
        \begin{tikzpicture}[font=\footnotesize]
          \node at (0.00,0.51) {$R_{x}$};
          \node at (0.86,0.51) {$R_{y}$};
          \node at (1.72,0.51) {$R_{z}$};
          \node at (2.58,0.51) {$L_{x}$};
          \node at (3.44,0.51) {$L_{y}$};
          \node at (4.30,0.51) {$L_{z}$};
          \node[anchor=east] at (-0.50,0.00) {$R_{x}$};
          \node[anchor=east] at (-0.50,-0.55) {$R_{y}$};
          \node[anchor=east] at (-0.50,-1.10) {$R_{z}$};
          \node[anchor=east] at (-0.50,-1.65) {$L_{x}$};
          \node[anchor=east] at (-0.50,-2.20) {$L_{y}$};
          \node[anchor=east] at (-0.50,-2.75) {$L_{z}$};
          \node[fill=black!5, minimum width=0.86cm, minimum height=0.55cm, inner sep=0pt, text=black!40] at (0.00,0.00) {--};
          \node[fill=tehot!97, minimum width=0.86cm, minimum height=0.55cm, inner sep=0pt, text=white] at (0.86,0.00) {1.03};
          \node[fill=tehot!35, minimum width=0.86cm, minimum height=0.55cm, inner sep=0pt, text=black] at (1.72,0.00) {0.37};
          \node[fill=tehot!43, minimum width=0.86cm, minimum height=0.55cm, inner sep=0pt, text=black] at (2.58,0.00) {0.46};
          \node[fill=tehot!39, minimum width=0.86cm, minimum height=0.55cm, inner sep=0pt, text=black] at (3.44,0.00) {0.42};
          \node[fill=tehot!41, minimum width=0.86cm, minimum height=0.55cm, inner sep=0pt, text=black] at (4.30,0.00) {0.43};
          \node[fill=tehot!97, minimum width=0.86cm, minimum height=0.55cm, inner sep=0pt, text=white] at (0.00,-0.55) {1.03};
          \node[fill=black!5, minimum width=0.86cm, minimum height=0.55cm, inner sep=0pt, text=black!40] at (0.86,-0.55) {--};
          \node[fill=tehot!30, minimum width=0.86cm, minimum height=0.55cm, inner sep=0pt, text=black] at (1.72,-0.55) {0.32};
          \node[fill=tehot!40, minimum width=0.86cm, minimum height=0.55cm, inner sep=0pt, text=black] at (2.58,-0.55) {0.43};
          \node[fill=tehot!37, minimum width=0.86cm, minimum height=0.55cm, inner sep=0pt, text=black] at (3.44,-0.55) {0.39};
          \node[fill=tehot!37, minimum width=0.86cm, minimum height=0.55cm, inner sep=0pt, text=black] at (4.30,-0.55) {0.40};
          \node[fill=tehot!26, minimum width=0.86cm, minimum height=0.55cm, inner sep=0pt, text=black] at (0.00,-1.10) {0.28};
          \node[fill=tehot!21, minimum width=0.86cm, minimum height=0.55cm, inner sep=0pt, text=black] at (0.86,-1.10) {0.22};
          \node[fill=black!5, minimum width=0.86cm, minimum height=0.55cm, inner sep=0pt, text=black!40] at (1.72,-1.10) {--};
          \node[fill=tehot!17, minimum width=0.86cm, minimum height=0.55cm, inner sep=0pt, text=black] at (2.58,-1.10) {0.19};
          \node[fill=tehot!17, minimum width=0.86cm, minimum height=0.55cm, inner sep=0pt, text=black] at (3.44,-1.10) {0.18};
          \node[fill=tehot!17, minimum width=0.86cm, minimum height=0.55cm, inner sep=0pt, text=black] at (4.30,-1.10) {0.18};
          \node[fill=tehot!37, minimum width=0.86cm, minimum height=0.55cm, inner sep=0pt, text=black] at (0.00,-1.65) {0.39};
          \node[fill=tehot!37, minimum width=0.86cm, minimum height=0.55cm, inner sep=0pt, text=black] at (0.86,-1.65) {0.39};
          \node[fill=tehot!11, minimum width=0.86cm, minimum height=0.55cm, inner sep=0pt, text=black] at (1.72,-1.65) {0.11};
          \node[fill=black!5, minimum width=0.86cm, minimum height=0.55cm, inner sep=0pt, text=black!40] at (2.58,-1.65) {--};
          \node[fill=tehot!78, minimum width=0.86cm, minimum height=0.55cm, inner sep=0pt, text=white] at (3.44,-1.65) {0.83};
          \node[fill=tehot!69, minimum width=0.86cm, minimum height=0.55cm, inner sep=0pt, text=white] at (4.30,-1.65) {0.73};
          \node[fill=tehot!40, minimum width=0.86cm, minimum height=0.55cm, inner sep=0pt, text=black] at (0.00,-2.20) {0.43};
          \node[fill=tehot!41, minimum width=0.86cm, minimum height=0.55cm, inner sep=0pt, text=black] at (0.86,-2.20) {0.44};
          \node[fill=tehot!11, minimum width=0.86cm, minimum height=0.55cm, inner sep=0pt, text=black] at (1.72,-2.20) {0.12};
          \node[fill=tehot!71, minimum width=0.86cm, minimum height=0.55cm, inner sep=0pt, text=white] at (2.58,-2.20) {0.75};
          \node[fill=black!5, minimum width=0.86cm, minimum height=0.55cm, inner sep=0pt, text=black!40] at (3.44,-2.20) {--};
          \node[fill=tehot!74, minimum width=0.86cm, minimum height=0.55cm, inner sep=0pt, text=white] at (4.30,-2.20) {0.79};
          \node[fill=tehot!46, minimum width=0.86cm, minimum height=0.55cm, inner sep=0pt, text=black] at (0.00,-2.75) {0.49};
          \node[fill=tehot!47, minimum width=0.86cm, minimum height=0.55cm, inner sep=0pt, text=black] at (0.86,-2.75) {0.50};
          \node[fill=tehot!14, minimum width=0.86cm, minimum height=0.55cm, inner sep=0pt, text=black] at (1.72,-2.75) {0.15};
          \node[fill=tehot!100, minimum width=0.86cm, minimum height=0.55cm, inner sep=0pt, text=white] at (2.58,-2.75) {1.07};
          \node[fill=tehot!97, minimum width=0.86cm, minimum height=0.55cm, inner sep=0pt, text=white] at (3.44,-2.75) {1.03};
          \node[fill=black!5, minimum width=0.86cm, minimum height=0.55cm, inner sep=0pt, text=black!40] at (4.30,-2.75) {--};
          \draw[thick] (-0.43,0.28) rectangle (2.15,-1.38);
          \draw[thick] (2.15,-1.38) rectangle (4.73,-3.03);
          \node[anchor=north] at (2.15,-3.19) {Source node};
          \node[rotate=90, anchor=south] at (-1.39,-1.38) {Target node};
        \end{tikzpicture}
    \par\vspace{1.6em}
    \begin{tikzpicture}[
        every node/.style={font=\scriptsize, inner sep=0pt, anchor=west},
        core_leg/.style={circle, draw, thick, fill=corered!35, minimum size=3.2mm, anchor=center},
        trueedge/.style={-{Latex}, semithick},
        spurious/.style={-{Latex}, dotted, semithick}
    ]
        \node[core_leg] at (0,0) {};
        \node at (0.34,0) {core node};
        \draw[trueedge] (-0.16,-0.46) -- (0.16,-0.46);
        \node at (0.34,-0.46) {true causal neighbor};
        \draw[spurious] (-0.16,-0.92) -- (0.16,-0.92);
        \node at (0.34,-0.92) {spurious connection};
    \end{tikzpicture}
\end{minipage}\hfill
\begin{minipage}[t]{0.26\linewidth}
    \centering
    {\bfseries b}\par\vspace{0.5em}
    \drawsystem{R\"ossler}{Rx}{Ry/Rx}{}\par\vspace{0.2em}
    {\scriptsize $q=1$}\par\vspace{0.9em}
    \drawsystem{R\"ossler}{Rx}{Ry/Rx, Rz/Rx}{Lx/Rx, Ly/Rx, Lz/Rx}\par\vspace{0.2em}
    {\scriptsize $q=5$}
\end{minipage}\hfill
\begin{minipage}[t]{0.26\linewidth}
    \centering
    {\bfseries c}\par\vspace{0.5em}
    \drawsystem{R\"ossler}{Rz}{Rx/Rz}{}\par\vspace{0.2em}
    {\scriptsize core node $R_z$}\par\vspace{0.9em}
    \drawsystem{R\"ossler}{Lx}{Ly/Lx, Lz/Lx}{Rx/Lx, Ry/Lx}\par\vspace{0.2em}
    {\scriptsize core node $L_x$, threshold $=0.28$}
\end{minipage}

\caption{No global hyperparameter recovers the neighborhoods of the composite Lorenz63--R\"ossler system.
\textbf{a}, The \acf{TE} matrix of a single attractor pair, from which both reconstructions below are read off. Entry $(i,j)$ is the \ac{TE} from source node $j$ to target node $i$, in the convention of Fig.~\ref{fig:lorenz_roessler}c of the main text, which shows the same matrix without values; cell shading is proportional to the entry, and the black outlines mark the two ground-truth blocks. Diagonal entries are not scored.
\textbf{b}, Global neighborhood size: with $q=1$ the true neighbor $R_z$ of the core node $R_x$ is missed; with $q=5$, the smallest size that includes $R_z$, the three independent Lorenz variables are admitted as well, since \ac{TE} ranks them above $R_z$. No $q$ recovers the neighborhood of $R_x$ correctly.
\textbf{c}, Global threshold: $0.28$ is the highest threshold that still captures the true neighbor $R_x$ of the core node $R_z$ (top); at this value the neighborhood of $L_x$ (bottom) already contains spurious R\"ossler variables. No single threshold recovers both neighborhoods without introducing spurious edges in one or false negatives in the other.}
\label{fig:si_global_hyperparameter}
\end{figure}
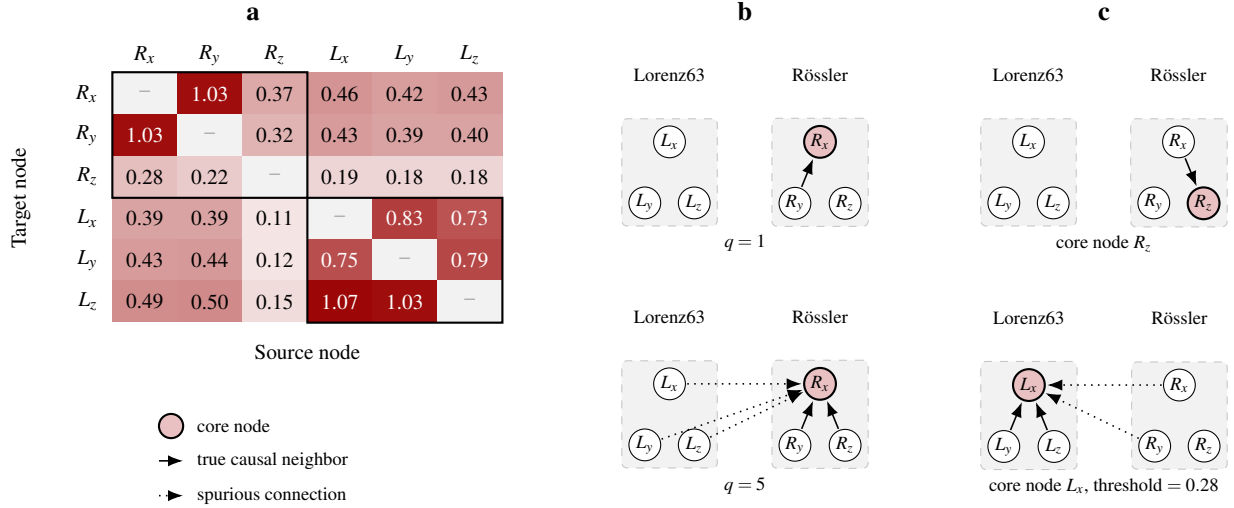

The analysis is based on the \acf{TE} matrix of a single attractor pair, reproduced with its values in Fig.~\ref{fig:si_global_hyperparameter}a; it is the same matrix that is shown as a heatmap in Fig.~\ref{fig:lorenz_roessler}c of the main text. According to the governing equations, the parents of the R\"ossler coordinate $R_x$ are $R_y$ and $R_z$. \Ac{TE} recovers $R_y$ ($1.03$) but scores the second parent $R_z$ at only $0.37$, below all three of the entirely independent Lorenz variables ($L_x = 0.46$, $L_z = 0.43$, $L_y = 0.42$).
A global neighborhood size, therefore, cannot separate the attractors: $q=1$ misses the true neighbor $R_z$, while $q=5$, the smallest size that includes $R_z$, necessarily includes the three spurious Lorenz variables ranked above it (Fig.~\ref{fig:si_global_hyperparameter}b). No value of $q$ recovers the neighborhood of $R_x$ correctly. While such misclassifications may appear negligible in six dimensions, false positives accumulate as the system dimension grows, occupying an increasing fraction of each local model's feature vector, degrading prediction performance and obscuring the inferred structure.

A global threshold fails for a similar reason. The two attractors evolve on different scales, so \ac{TE} values are not comparable across the subsystems. The highest threshold that still captures the true neighbor $R_x$ of the core node $R_z$ is $0.28$, the value of that entry; at this threshold the neighborhood of $L_x$ already contains the spurious R\"ossler variables $R_y$ ($0.39$) and $R_x$ ($0.39$), both of which have a higher TE score than $R_z$ in the $R_x$ neighborhood (Fig.~\ref{fig:si_global_hyperparameter}c). Conversely, any global threshold high enough to ensure a correct neighborhood of $L_x$ produces false negatives for the neighborhood of $R_z$. No single threshold recovers both neighborhoods correctly.

\suppnote{\quad Why a network reconstruction with causal-discovery algorithms is not optimal for prediction}
\label{sec:supp_causal_vs_prediction}

Causal-discovery algorithms aim to recover the true causal graph of a system. They are designed and benchmarked for structural correctness rather than for forecast accuracy, and the structure best suited for prediction, especially from finite data, need not coincide with the graph they return. 

\paragraph{Causal discovery optimises structure recovery, not prediction.}
Constraint-based algorithms such as the PC algorithm~\citeSI{SI_Spirtes_CausationPredictionSearch} and its time-series extension PCMCI~\citeSI{SI_Runge2019_PCMCI} decide the presence of each edge with conditional-independence tests carried out at a single global significance level $\alpha$, applied uniformly to every variable. Their objective is to recover the true graph, and they are evaluated with structural metrics accounting for every misplaced edge equally, regardless of how much that edge matters for forecasting. 

\paragraph{Causal Markov Condition.}
One might object that the true causal graph is already optimal for prediction. This intuition rests on the causal Markov condition~\citeSI{SI_Neapolitan_probablistic_methods,SI_Runge_causalRecon2018}, which states that in a causal graph every variable $X$ is conditionally independent of its non-descendants $\mathrm{ND}(X)$ given its parents,
\begin{equation}
\label{eq:causal_markov}
X \;\perp\!\!\!\perp\; \bigl(\mathrm{ND}(X)\setminus \mathrm{Pa}(X)\bigr)\;\bigm|\;\mathrm{Pa}(X).
\end{equation}
It implies that the parents of a node are a sufficient input for predicting it, and it is what justifies decomposing a $d$-dimensional forecasting problem into $d$ local neighborhood problems. The true causal parents are therefore an optimal predictor set. But Eq.~\eqref{eq:causal_markov} is a statement about conditional independencies in the infinite-data limit. In a finite or real data setting, a causal discovery algorithm is not optimal for inferring a graph that facilitates prediction. First, the effects on the prediction quality differ greatly between edges. In a finite or real data setting, a causal discovery algorithm is expected to make mistakes. Trying to recover all edges with equal priority might therefore harm predictive performance, and an algorithm that takes predictive performance into account during the network reconstruction is preferable. Second, causal discovery algorithms try to specifically filter out hidden confounder effects, as they are not true causal links. However, while such a link is spurious, it can still carry usable predictive information. The next two paragraphs illustrate those points.

\begin{figure*}[t]
\centering
\includegraphics[width=\linewidth]{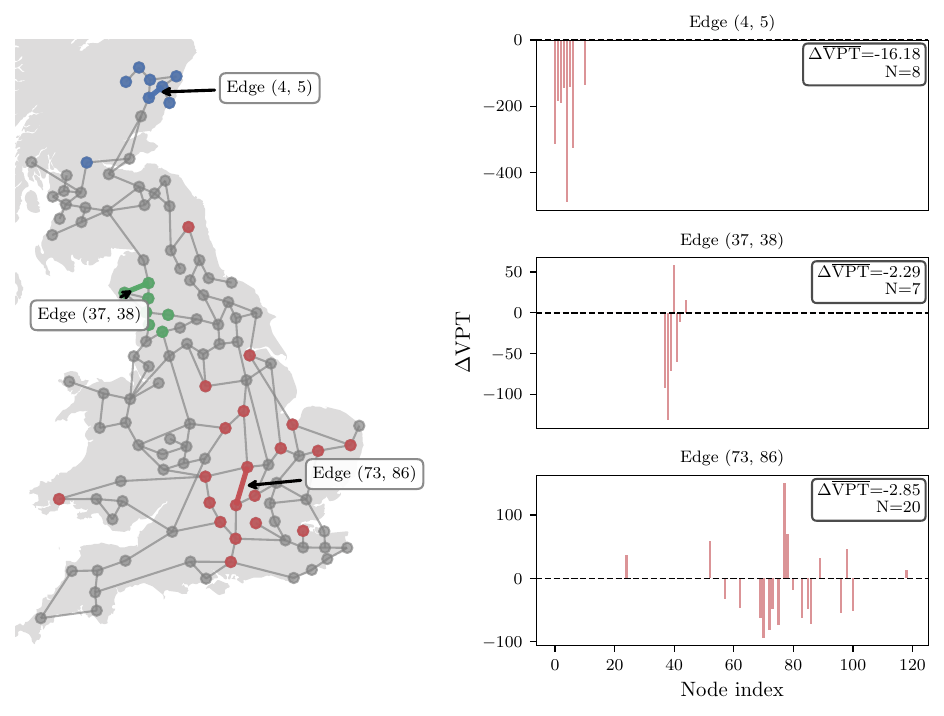}
\caption{Edge-importance experiment on the UK power grid with three representative single-edge removals shown. In each case, two reservoir-computing local-state models are trained, one with the true adjacency matrix and one with the selected edge removed; a node is counted as affected if its \acf{VPS} change by more than a threshold ($\mathrm{VPS_{th}}=5$). \textbf{Left:} the removed edges and the corresponding affected nodes highlighted on the UK power grid network. \textbf{Right:} node-wise change in \ac{VPS} for each removal. Some edges affect many nodes, others only a few.}
\label{fig:si_edge_importance_samples}
\end{figure*}

\begin{figure*}[t]
\centering
\includegraphics[width=\linewidth]{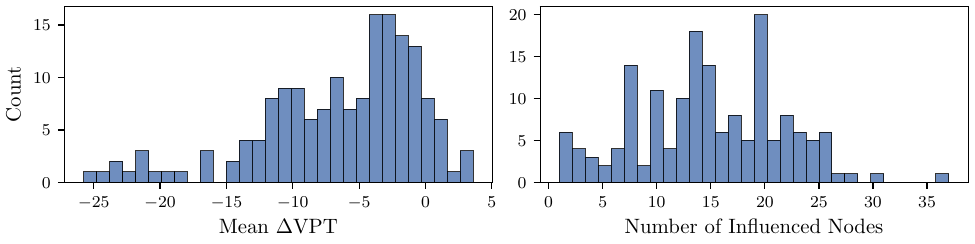}
\caption{Distribution of the change in \acf{VPS} and of the number of affected nodes across all removed edges. The importance of an edge for prediction is highly heterogeneous, so that a criterion aimed at structural recovery rather than forecast skill is suboptimal when the goal is prediction.}
\label{fig:si_edge_importance_statistics}
\end{figure*}

\paragraph{Importance of edges for prediction.}
This paragraph briefly answers the question if all edges are of equal importance for prediction. For this experiment we trained two reservoir-computing local-states models, one supplied with the true network and the other with one selected edge missing from the true network. Then the change in the per-node \acf{VPS} relative to the full-network baseline is recorded (a node counts as affected if its \ac{VPS} changes by more than five steps). Figure~\ref{fig:si_edge_importance_samples} shows three representative removals. Removing a single edge can sharply degrade the forecast of the nodes around it, and different edges affect very different numbers of nodes and by different degrees. The summary statistics across the full UK power grid are provided in Figure~\ref{fig:si_edge_importance_statistics}, which confirms that predictive importance seems to vary widely across edges. Another instructive case arises in systems whose nodes can synchronize. Consider a cluster of neighbors of a core node $i$ whose states are (generalized-)synchronized such that each of these nodes is a smooth function of any other, $X_k(t)=\psi_{kj}(X_j(t))$. Including one representative of the cluster in the neighborhood of $i$ can be highly valuable, as it captures the cluster's entire influence on the core node. Every further member, however, is (approximately) a deterministic function of a variable already conditioned on, so its additional predictive information vanishes. From the standpoint of network reconstruction all cluster edges are equally real and equally worth recovering, whereas for prediction only one of them matters. An algorithm that selects neighborhoods by forecast skill naturally exploits this redundancy structure, whereas classical causal discovery cannot.

\begin{figure}[t]
\centering
\begin{tikzpicture}[
    scale=0.85,
    every node/.style={circle, draw, thick, minimum size=1.1cm},
]
\node (H) at (0,0) {$H$};
\node (X) at (3,0) {$X$};
\node (Y) at (3,-2.2) {$Y$};

\draw[->, thick]
    (H) edge[bend left=15]
    node[midway, above, draw=none, yshift=-6pt] {$\tau=1$} (X);

\draw[->, thick]
    (H) edge[bend right=15]
    node[midway, left, draw=none, yshift=-6pt, xshift=4pt] {$\tau=2$} (Y);

\draw[->, thick, dotted]
    (X) -- (Y);
\end{tikzpicture}
\caption{Hidden confounder inducing a spurious but predictive link. The hidden process $H$ drives $X$ at lag $\tau=1$ and $Y$ at lag $\tau=2$. There is no direct causal link $X\to Y$ (dotted), yet $X$ is predictive of the future of $Y$ because both are echoes of $H$. Consequently a spurious $X\to Y$ link improves the forecast of $Y$.}
\label{fig:si_spurious_link}
\end{figure}
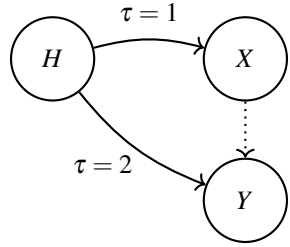

\paragraph{Spurious links can improve prediction.}
A hidden confounder is an unmeasured cause of two or more observed variables, which might introduce spurious links into the inferred network. While the assumption of causal sufficiency would prevent such a situation, it is rarely possible to observe all variables relevant to a real system \citeSI{SI_Runge_causalRecon2018}.
A minimal example makes this precise (Fig.~\ref{fig:si_spurious_link}). Let $H_t$ be a hidden process evolving independently, $H_t \sim \mathcal{N}(0,1)$, and let two observed variables be generated as
\begin{align}
X_t &= H_{t-1} + \epsilon^X_t, \\
Y_t &= H_{t-2} + \epsilon^Y_t,
\end{align}
with independent noise $\epsilon^X_t,\epsilon^Y_t \sim \mathcal{N}(0,\sigma^2)$. The only true causal links are $H_{t-1}\to X_t$ and $H_{t-2}\to Y_t$; there is no direct influence of $X$ on $Y$. Nevertheless, because $X_t$ and $Y_{t+1}$ are both echoes of the same hidden state ($X_t$ reflects $H_{t-1}$, and so does $Y_{t+1}=H_{t-1}+\epsilon^Y_{t+1}$), the two are statistically dependent, $\mathbb{E}[Y_{t+1}\mid X_t]\neq\mathbb{E}[Y_{t+1}]$. Using $X$ to predict $Y$ therefore improves the forecast even though the link $X\to Y$ is spurious. A causal-discovery algorithm that correctly reports the absence of a direct $X\to Y$ edge would discard exactly the information that helps predict $Y$. The same effect arises even without hidden variables, through Takens' embedding theorem, and is visible in the Lorenz96 benchmark of the main text, where \acf{CLS} retains nearby non-parent nodes that improve the forecast (Fig.~\ref{fig:lorenz96}).

\suppnote{Choice of the wrapper model and self-influence}
\label{sec:supp_wrapper}

The wrapper that is used to score candidate neighborhoods can in principle be any predictive model. We compare two different established models: \acf{RC} and \acf{NGRC}. \Ac{NGRC} reaches comparable accuracy at lower training cost and with fewer hyperparameters, which already suits the many wrapper evaluations \ac{CLS} performs. It is not immediately clear how a wrapper model should be used to quantify neighborhood quality. Depending on the strategy, the time-delayed values of the core node itself are included or excluded from the neighborhood~(which~we~call~self-influence). 

We compare three sample strategies:
\begin{itemize}
\item \textbf{One-step (\ac{RC} or \ac{NGRC}).} The full neighborhood is reset to the true values at every step and the model makes a one-step forecast of the core node.
\item \textbf{Multi-step (\ac{RC} or \ac{NGRC}).} The neighborhood is reset only every $N_{\mathrm{reset}}$ steps, so the forecast error is allowed to accumulate between resets.
\item \textbf{\Ac{NGRC} inference mode.} \Ac{NGRC} is used in inference mode, which learns a map from the neighbors alone to the next core state and therefore excludes the core node's own past from the feature vector. This means that there is no self-influence in this strategy.
\end{itemize}

\begin{figure*}[t]
\centering
\includegraphics[width=\linewidth]{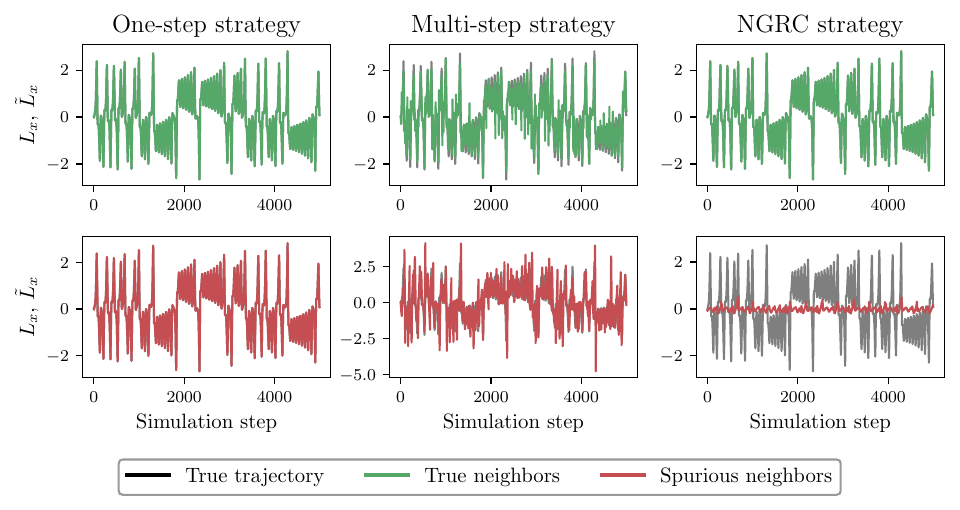}
\caption{Sample core-node predictions for a composite Lorenz63--R\"ossler system. For the core node $L_x$, wrapper models are trained on a neighborhood of the true neighbors $L_y,L_z$ (green) and on a neighborhood of spurious neighbors $R_x,R_y,R_z$ (red), for each of the three strategies. Under the one-step strategy the core node's own past dominates, so even the spurious neighborhood reproduces the trajectory; under \acf{NGRC} inference mode the spurious neighborhood fails to track the core trajectory at all.}
\label{fig:si_wrapper_traj}
\end{figure*}

\begin{figure*}[t]
\centering
\subfloat[One-step strategy]{\includegraphics[width=0.72\linewidth]{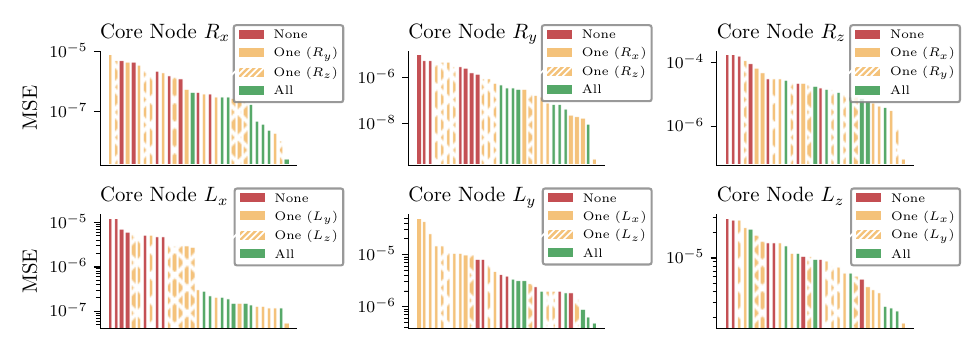}}\\
\subfloat[Multi-step strategy]{\includegraphics[width=0.72\linewidth]{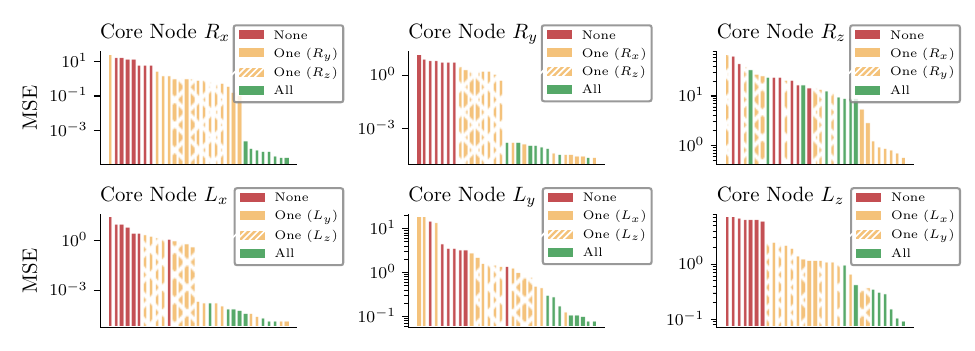}}\\
\subfloat[NGRC inference mode]{\includegraphics[width=0.72\linewidth]{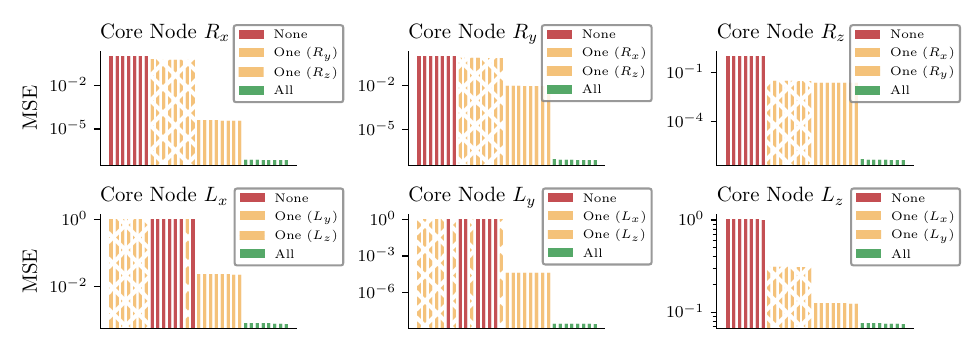}}
\caption{Core-node \acf{MSE} for all $32$ candidate neighborhoods of the composite Lorenz63--R\"ossler system, evaluated for each core node and grouped by how many features come from the same attractor as the core node: none (red), one (orange, the two possible cases distinguished), or both (green). Only \acf{NGRC} inference mode \textbf{(c)} cleanly separates the neighborhoods that contain the true neighbors from those that do not; the one-step strategy \textbf{(a)} barely separates them because the core node dominates the forecast, and the multi-step strategy \textbf{(b)} improves the separation but does not fully resolve it.}
\label{fig:si_wrapper_mse}
\end{figure*}

A good wrapper should assign a clearly lower error to neighborhoods that contain the true neighbors than to those built from spurious ones. Figure~\ref{fig:si_wrapper_traj} shows that only \ac{NGRC} inference mode achieves this cleanly.
Our hypothesis is that because in the one-step strategy the core node is part of the feature vector, and because a variable's own past is usually strongly predictive of its next value, the core node dominates the forecast and masks the contribution of the neighbors. The multi-step strategy mitigates the masking slightly by letting the error accumulate between resets, which improves the separation. 
To confirm this, we can look at two representative sample predictions. Fig.~\ref{fig:si_wrapper_mse} shows the \acf{MSE} of the core node predictions for two different cases: (i) all true neighbors are present in the neighborhood, which is highlighted in green, (ii) no true neighbors are present in the neighborhood, highlighted in red. For a functional wrapper, the prediction on the spurious neighborhood should be nonsensical. 
As evident from this figure, only the \ac{NGRC} inference strategy, which has no self-influence, is successful in this test. 

Importantly, while excluding the self-influence during the graph reconstruction is desirable, we, nevertheless, \textit{do} include the core node time delays as local \ac{NGRC} input during the full system forecast, as at this stage the information about the recent past of the core node becomes important for prediction and the masking effect is no longer relevant. 
The one-step strategy is also expensive, since resetting the neighborhood at every step forces the reservoir to resynchronize before every prediction. The multi-step strategy reduces the number of re-synchronizations (Fig.~\ref{fig:si_wrapper_mse}b), but is still slower than the \ac{NGRC} inference mode.

These considerations lead us to use \ac{NGRC} in inference mode as the wrapper throughout this paper and \ac{NGRC} as the per-node model in the full-system forecast. However, this choice is not fundamental, as the framework stays model-agnostic and \ac{NGRC} can be swapped out for other prediction models. 

\suppnote{Simulation and model hyperparameters}
\label{sec:supp_hyperparameters}

For reproducibility, Tables~\ref{tab:supp_l63r}--\ref{tab:supp_ukpg} collect the full simulation and model hyperparameters used to produce the three benchmark results reported in the main text. 
For each benchmark we list the parameters of the dynamical system (whose time series data are obtained by integration with the Runge-Kutta method RK4), of the causal measure used in the filter step (\ac{TE} or \acf{CCM}), of the wrapper and prediction models (\ac{NGRC} and its variants), and of the \ac{CLS} neighborhood-selection thresholds $\alpha_1$ (wrapper selection) and $\alpha_2$ (backward elimination).
The benchmark systems are deterministic, so the random seed affects only the choice of the initial condition. For the UK power grid the seed sets the natural frequencies $\omega_i$ and the initial phases.

During the wrapper step, the training data is split up further into a wrapper train and a wrapper test set. This is to ensure that the actual test set used to evaluate the final prediction is unseen during the wrapper step. We denote this split with the wrapper test fraction, which just identifies how much of the original training data is used to evaluate the wrapper performance.

As the objective of the paper is not to showcase the highest performance possible on the benchmarks, there was no deep hyperparameter search conducted. The hyperparameters were handpicked such that the model produced interpretable results (e.g. no divergence of the \ac{NGRC} prediction).

\begin{table*}[t]
\centering
\footnotesize
\begin{tabular}{@{}lccc@{}}
\toprule
 & \textbf{Exhaustive wrapper} & \textbf{Single attractor pair} & \textbf{Scaling sweep} \\
 & \textbf{validation (Fig.~2b)} & \textbf{(Fig.~2c--e)} & \textbf{(Fig.~2f--h)} \\
\midrule
\multicolumn{4}{@{}l}{\textbf{System}} \\
Attractor pairs $N$ & $1$ & $1$ & $1$--$15$ \\
Lorenz63 $\mathrm{d}t$; $\sigma,\rho,\beta$ & \multicolumn{3}{c}{$0.01$; \ \ $10.0,\ 28.0,\ 8/3$} \\
R\"ossler $\mathrm{d}t$; $a,b,c$ & \multicolumn{3}{c}{$0.05$; \ \ $0.2,\ 0.2,\ 5.7$} \\
Initial condition, every subsystem & \multicolumn{3}{c}{drawn from $\mathcal{U}(-3,3)^3$ with the seed below} \\
Random seed & $42$ & $42$ & $0,1,\dots,9$ (ten seeds) \\
\midrule
\multicolumn{4}{@{}l}{\textbf{Data}} \\
Transient steps (discarded) & \multicolumn{3}{c}{$\ \ \ 30000$} \\
Data length (post-transient) & $50000$ & $50000$ & $56050$ \\
Train / test steps & $25000\ /\ 25000$ & $25000\ /\ 25000$ & $25000\ /\ 31050$ \\
Normalization & \multicolumn{3}{c}{zero mean, unit variance} \\
\midrule
\multicolumn{4}{@{}l}{\textbf{Transfer entropy (filter)}} \\
Number of bins & --- & $79$ & $8$ \\
Time delay $\tau$ & --- & $1$ & $1$ \\
\midrule
\multicolumn{4}{@{}l}{\textbf{NGRC (wrapper \& local states)}} \\
Delays $k$, spacing $s$ & $5,\ 1$ & $3,\ 1$ & $3,\ 1$ \\
Polynomial orders & $[1,2]$ & $[1,2,3]$ & $[1,2,3]$ \\
Ridge regularisation & $10^{-2}$ & $10^{-3}$ & $10^{-3}$ \\
Wrapper test fraction & $0.2$ & $0.4$ & $0.4$ \\
Bias term & \multicolumn{3}{c}{$ \ \ $ yes} \\
\midrule
\multicolumn{4}{@{}l}{\textbf{CLS selection}} \\
$\alpha_1,\ \alpha_2$ & --- & $0.05,\ 0.05$ & $0.05,\ 0.05$ \\
Candidate bound $d_{\max}$ & --- & unbounded ($=d-1=5$) & $10$ \\
\midrule
\multicolumn{4}{@{}l}{\textbf{Forecast and evaluation}} \\
Forecast horizon $N_{\mathrm{pred}}$ & --- & $5000$ & $3000$ \\
VPT launch points $\times$ warmup & --- & \multicolumn{2}{c}{$10\ \times\ 50$ steps} \\
VPT threshold $\epsilon$ & --- & $0.3$ & $0.3$ \\
Monolithic NGRC baseline & --- & $k{=}3,s{=}1$, orders $[1,2,3]$ & $k{=}3,s{=}1$, orders $[1]$ \\
 & & ridge $10^{-3}$ & ridge $10^{-3}$ \\
\bottomrule
\end{tabular}
\caption{Simulation and model hyperparameters for the composite Lorenz63--R\"ossler benchmark. \acs{NGRC}, \acl{NGRC}; \acs{CLS}, \acl{CLS}; \acs{VPT}, \acl{VPT}.}
\label{tab:supp_l63r}
\end{table*}
\clearpage
\begin{table}[H]
\centering
\footnotesize
\begin{tabular}{@{}ll@{}}
\toprule
\multicolumn{2}{@{}l}{\textbf{Lorenz96 system}} \\
Dimension $N$ & $40$ \\
Forcing $F$ & $5$ \\
Time step $\mathrm{d}t$ & $0.05$ \\
Initial condition & $x_0 = F\cdot\mathbf{1} + 10^{-5}\cdot\mathcal{U}(0,1)^N$ \\
Random seeds & $0,1,\dots,9$ (ten seeds) \\
Transient steps (discarded) & $10000$ \\
Train / test & $20000\ /\ 10150\ $ \\
Normalization & zero mean, unit variance \\
\midrule
\multicolumn{2}{@{}l}{\textbf{Transfer entropy (filter)}} \\
Number of bins & $30$ \\
Time delay $\tau$ & $1$ \\
Data used for TE estimation & train split ($20000$ steps) \\
\midrule
\multicolumn{2}{@{}l}{\textbf{NGRC (wrapper: selection \& elimination)}} \\
Delays $k$, spacing $s$ & $3,\ 1$ \\
Polynomial orders & $[1,2]$ \\
Wrapper test fraction & $0.3$ \\
Ridge regularisation & $10^{-4}$ \\
\midrule
\multicolumn{2}{@{}l}{\textbf{NGRC (local states)}} \\
Delays $k$, spacing $s$ & $5,\ 1$ \\
Polynomial orders & $[1,2]$ \\
Ridge regularisation & $10^{-1}$ \\
Bias term & yes \\
\midrule
\multicolumn{2}{@{}l}{\textbf{CLS selection}} \\
$\alpha_1,\ \alpha_2$ & $0.03,\ 0.03$ \\
Candidate bound $d_{\max}$ & $20$ \\
\midrule
\multicolumn{2}{@{}l}{\textbf{Evaluation}} \\
VPT launch points $\times$ warmup & $10\ \times\ 50$ steps \\
VPT threshold $\epsilon$ & $0.3$ \\
\bottomrule
\end{tabular}
\caption{Simulation and model hyperparameters for the Lorenz96 benchmark. \acs{TE}, \acl{TE}; \acs{NGRC}, \acl{NGRC}; \acs{CLS}, \acl{CLS}; \acs{VPT}, \acl{VPT}.}
\label{tab:supp_l96}
\end{table}

\clearpage
\begin{table}[H]
\centering
\footnotesize
\begin{tabular}{@{}lcc@{}}
\toprule
 & \textbf{Sine-NGRC} & \textbf{Kuramoto-NGRC} \\
\midrule
\multicolumn{3}{@{}l}{\textbf{UK power grid}} \\
Phase lags $\alpha,\ \beta$ & \multicolumn{2}{c}{$0.02,\ 0.06$} \\
Coupling $\gamma_1,\ \gamma_2$ & \multicolumn{2}{c}{$0.4,\ 0.4$} \\
$\omega_{\mathrm{abs,low}}$ & \multicolumn{2}{c}{$0.3$} \\
Time step $\mathrm{d}t$ & \multicolumn{2}{c}{$8\times10^{-2}$} \\
Initial phases & \multicolumn{2}{c}{$\mathcal{U}(0,2\pi)^N$, drawn with the seed} \\
Random seeds & \multicolumn{2}{c}{$0,1,\dots,9$ (ten seeds)} \\
Transient steps (discarded) & \multicolumn{2}{c}{$1000$} \\
Train / test / forecast steps & \multicolumn{2}{c}{$10000\ /\ 20150\ /\ 2000$} \\
Normalization & \multicolumn{2}{c}{none (raw phases)} \\
\midrule
\multicolumn{3}{@{}l}{\textbf{Convergent cross mapping}} \\
Input series & \multicolumn{2}{c}{$\sin\theta_i$} \\
Lag & \multicolumn{2}{c}{$1$} \\
Embedding dimension & \multicolumn{2}{c}{$2$} \\
Training fraction & \multicolumn{2}{c}{$0.75$} \\
Library size & \multicolumn{2}{c}{$L_{\mathrm{train}}$ (largest library only)} \\
\midrule
\multicolumn{3}{@{}l}{\textbf{Wrapper and forecast models}} \\
Input series & \multicolumn{2}{c}{raw phases $\theta_i$} \\
Regression target & \multicolumn{2}{c}{$\Delta\theta_t = \theta_{t+1}-\theta_t$} \\
Nonlinear features & polynomials of $\sin\theta_i,\cos\theta_i$ & $\sin(\theta_j-\theta_i)$ \\
Delays $k$, spacing $s$ & $3,\ 1$ & $1,\ 1$ \\
Polynomial orders & $[1,2]$ & --- \\
Ridge (wrapper) & $10^{-3}$ & $10^{-3}$ \\
Ridge (forecast) & $10^{-3}$ & $10^{-2}$ \\
Wrapper test fraction & $0.4$ & $0.3$ \\
Bias term & yes & yes \\
\midrule
\multicolumn{3}{@{}l}{\textbf{CLS selection}} \\
$\alpha_1,\ \alpha_2$ & $0.05,\ 0.05$ & $0.05,\ 0.05$ \\
Candidate bound $d_{\max}$ & $10$ & $30$ \\
Backward elimination & greedy & greedy \\
\midrule
\multicolumn{3}{@{}l}{\textbf{Evaluation}} \\
VPT threshold $\epsilon$ & \multicolumn{2}{c}{$0.3$} \\
VPS / VPT evaluated on & \multicolumn{2}{c}{$\sin\theta_i$} \\
\bottomrule
\end{tabular}
\caption{Simulation and model hyperparameters for the UK power grid benchmark. The seed fixes both the natural frequencies $\omega_i$ and the initial phases. \acs{NGRC}, \acl{NGRC}; \acs{CLS}, \acl{CLS}; \acs{VPS}, \acl{VPS}; \acs{VPT}, \acl{VPT}.}
\label{tab:supp_ukpg}
\end{table}

\bibliographystyleSI{naturemag-doi}
\bibliographySI{SI_references}

\end{document}